\documentclass[10pt,twocolumn,letterpaper]{article}

\usepackage[pagenumbers]{iccv} 
\newcommand{\bs}{\boldsymbol}

\DeclareMathOperator{\argmin}{argmin}

\definecolor{iccvblue}{rgb}{0.21,0.49,0.74}
\definecolor{ddt}{HTML}{38a3a5}

\usepackage{fontawesome5} 
\usepackage{amsthm}
\usepackage{multirow}
\usepackage[pagebackref,breaklinks,colorlinks,allcolors=iccvblue,urlcolor=ddt]{hyperref}
\newtheorem{lemma}{Lemma}

\def\paperID{233333333}
\def\confName{ICCV}
\def\confYear{2025}

\title{{\color{ddt}DDT}: {\color{ddt}D}ecoupled {\color{ddt}D}iffusion {\color{ddt}T}ransformer}

\author{
Shuai Wang\textsuperscript{1} \quad \quad Zhi Tian\textsuperscript{2} \quad \quad Weilin Huang\textsuperscript{2} \quad \quad Limin Wang \textsuperscript{1, {\color{lightgray} \faEnvelope}} \\
$^1$Nanjing University \quad  $^2$ByteDance Seed Vision \\  [0.2cm]
{\bf \url{https://github.com/MCG-NJU/DDT}} 
}

\newcommand\blfootnote[1]{%
  \begingroup
  \renewcommand\thefootnote{}\footnote{#1}%
  \addtocounter{footnote}{-1}%
  \endgroup
}

\begin{document}

\twocolumn[{%
\renewcommand\twocolumn[1][]{#1}%
\maketitle
\vspace{-1em}
\includegraphics[width=0.99\linewidth]{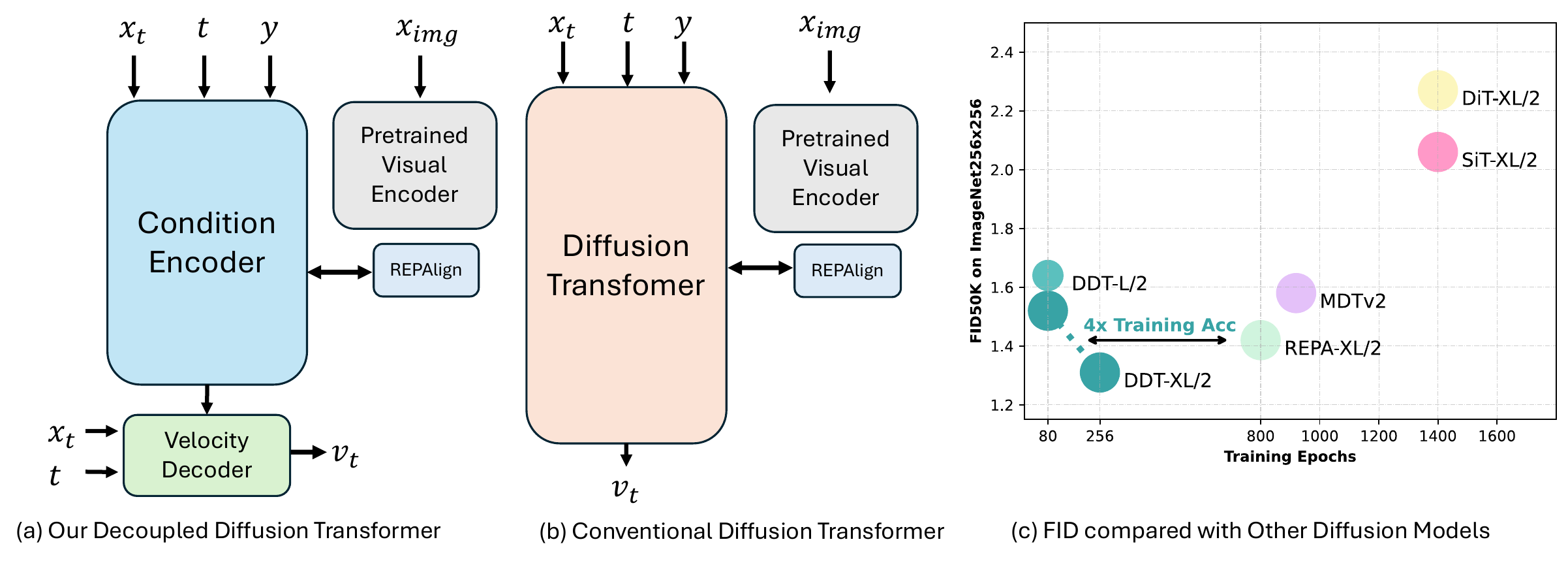}
\vspace{-1em}
\captionof{figure}{\textbf{Our deoupled diffusion transformer~(DDT-XL/2) achieves a SoTA 1.31 FID under 256 epochs.}~{\small Our decoupled diffusion transformer models incorporate a condition encoder to extract semantic self-conditions and a velocity decoder to decode velocity.}}
\label{fig:teaser}
\vspace{1em}
}]

\begin{abstract}
Diffusion transformers have demonstrated remarkable generation quality, albeit requiring longer training iterations and numerous inference steps. In each denoising step, diffusion transformers encode the noisy inputs to extract the lower-frequency semantic component and then decode the higher frequency with identical modules. This scheme creates an inherent optimization dilemma: encoding low-frequency semantics necessitates reducing high-frequency components, creating tension between semantic encoding and high-frequency decoding. To resolve this challenge, we propose a new \textbf{\color{ddt}D}ecoupled \textbf{\color{ddt}D}iffusion \textbf{\color{ddt}T}ransformer~(\textbf{\color{ddt}DDT}), with a decoupled design of a dedicated condition encoder for semantic extraction alongside a specialized velocity decoder. Our experiments reveal that a more substantial encoder yields performance improvements as model size increases. For ImageNet $256\times256$, Our DDT-XL/2 achieves a new state-of-the-art performance of {1.31 FID}~(nearly $4\times$ faster training convergence compared to previous diffusion transformers). For ImageNet $512\times512$, Our DDT-XL/2 achieves a new state-of-the-art FID of 1.28. Additionally, as a beneficial by-product, our decoupled architecture enhances inference speed by enabling the sharing self-condition between adjacent denoising steps. To minimize performance degradation, we propose a novel statistical dynamic programming approach to identify optimal sharing strategies. 

\end{abstract}  
\blfootnote{{\color{lightgray} \faEnvelope} \ : Corresponding author (lmwang@nju.edu.cn).}
\section{Introduction}
Image generation is a fundamental task in computer vision research, which aims at capturing the inherent data distribution of original image datasets and generating high-quality synthetic images through distribution sampling. Diffusion models~\cite{ddpm, vp, edm, flow, flow2} have recently emerged as highly promising solutions to learn the underlying data distribution in image generation, outperforming the GAN-based models~\cite{largegan, styleganxl} and Auto-Regressive models~\cite{maskgit, llamagen, rar}. 

The diffusion forward process gradually adds Gaussian noise to the pristine data following an SDE forward schedule~\cite{ddpm, vp, edm}. The denoising process learns the score estimation from this corruption process. Once the score function is accurately learned, data samples can be synthesized by numerically solving the reverse SDE~\cite{vp, edm, flow, flow2}.

Diffusion Transformers~\cite{dit, sit} introduce the transformer architecture into diffusion models to replace the traditionally dominant UNet-based model~\cite{uvit, adm}. Empirical evidence suggests that, given sufficient training iterations, diffusion transformers outperform conventional approaches even without relying on long residual connections~\cite{dit}. Nevertheless, their slow convergence rate still poses great challenge for developing new models due to the high cost.

In this paper, we want to tackle the aforementioned major disadvantages from a model design perspective. Classic computer vision algorithms~\cite{detr, sam, mae} strategically employ encoder-decoder architectures, prioritizing large encoders for rich feature extraction and lightweight decoders for efficient inference, while contemporary diffusion models predominantly rely on conventional decoder-only structures. We systematically investigate the underexplored potential of decoupled encoder-decoder designs in diffusion transformers, by answering the question of
\emph{{\color{ddt}can decoupled encoder-decoder transformer unlock the capability of accelerated convergence and enhanced sample quality?}}

Through investigation experiments, we conclude that the plain diffusion transformer has an optimization dilemma between abstract structure information extraction and detailed appearance information recovery. Further, the diffusion transformer is limited in extracting semantic representation due to the raw pixel supervision~\cite{repa, dod, mar}.
To address this issue, we propose a new architecture to explicitly decouple low-frequency semantic encoding and high-frequency detailed decoding through a customized encoder-decoder design. We call this encoder-decoder diffusion transformer model as \textbf{\color{ddt}DDT}~(\textbf{\color{ddt}D}ecoupled \textbf{\color{ddt}D}iffusion \textbf{\color{ddt}T}ransformer). DDT incorporates a {\em condition encoder} to extract semantic self-condition features. The extracted self-condition is fed into a {\em velocity decoder} along with the noisy latent to regress the velocity field. To maintain the local consistency of self-condition features of adjacent steps, we employ direct supervision of representation alignment and indirect supervision from the velocity regression loss of the decoder.

In the ImageNet$256\times256$ dataset, using the traditional off-shelf VAE~\cite{ldm}, our decoupled diffusion transformer~(DDT-XL/2) model achieves the state-of-the-art performance of 1.31 FID with interval guidance under only 256 epochs, approximately $4\times$ training acceleration compared to REPA~\cite{repa}. In the ImageNet$512\times512$ dataset, our DDT-XL/2 model achieves 1.28 FID within 500K finetuning steps.

Furthermore, our DDT achieves strong local consistency on its self-condition feature from the encoder. This property can significantly boost the inference speed by sharing the self-condition between adjacent steps. We formulate the optimal encoder sharing strategy solving as a classic minimal sum path problem by minimizing the performance drop of sharing self-condition among adjacent steps. We propose a statistic dynamic programming approach to find the optimal encoder sharing strategy with negligible second-level time cost. Compared with the naive uniform sharing, our dynamic programming delivers a minimal FID drop. Our contributions are summarized as follows.
\begin{itemize}
    \item We propose a new decoupled diffusion transformer model, which consists of a condition encoder and a velocity decoder.
    \item We propose statistic dynamic programming to find the optimal self-condition sharing strategy to boost inference speed while keeping minimal performance down-gradation.
    \item In the ImageNet$256\times256$ dataset, using tradition SDf8d4 VAE, our decoupled diffusion transformer~(DDT-XL/2) model achieves the SoTA 1.31 FID with interval guidance under only 256 epochs, approximately $4\times$ training acceleration compared to REPA~\cite{repa}. 
    \item In the ImageNet$512\times512$ dataset, our DDT-XL/2 model achieves the SoTA 1.28 FID, outperforming all previous methods with a significant margin.
\end{itemize}

\section{Related Work}
\begin{figure*}
    \centering
    \includegraphics[width=\linewidth]{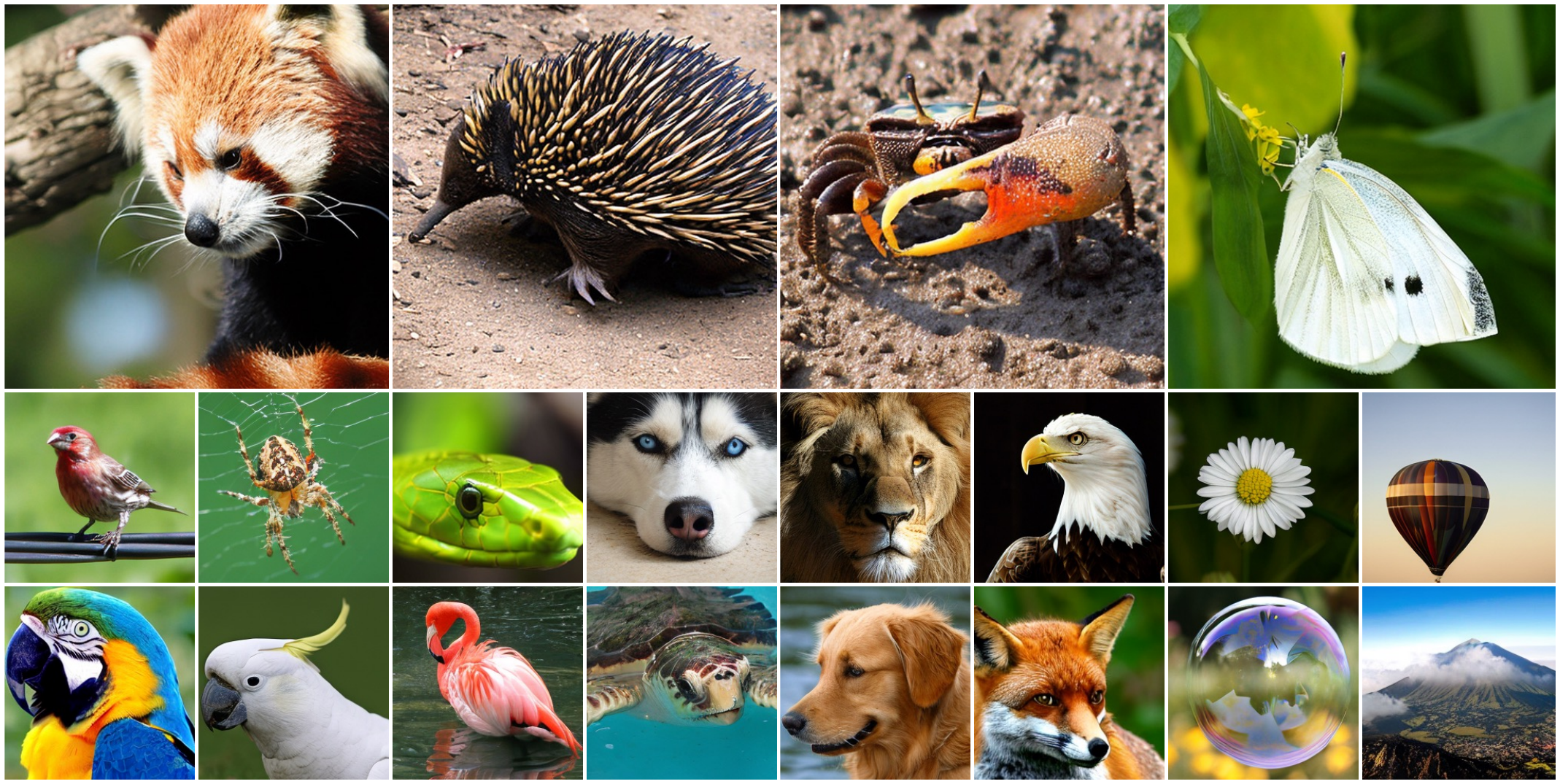}
    \caption{ \textbf{Selected $256\times256$ and $512\times512$ resolution samples.} {\small Generated from DDT-XL/2 trained on ImageNet  $256\times256$ resolution and ImageNet $512\times512$ resolution with CFG = 4.0.}}
    \vspace{-1em}
\end{figure*}

\paragraph{Diffusion Transformers.} The pioneering work of DiT~\cite{dit} introduced transformers into diffusion models to replace the traditionally dominant UNet architecture~\cite{uvit,adm}. Empirical evidence demonstrates that given sufficient training iterations, diffusion transformers outperform conventional approaches even without relying on long residual connections. SiT~\cite{sit} further validated the transformer architecture with linear flow diffusion. Following the simplicity and scalability of the diffusion transformer~\cite{sit,dit}, SD3~\cite{sd3}, Lumina~\cite{lumina}, and PixArt~\cite{pixart,pixart_sigma} introduced the diffusion transformer to more advanced text-to-image areas. Moreover, recently, diffusion transformers have dominated the text-to-video area with substantiated visual and motion quality~\cite{hunyuanvideo,cosmos,cogvideo}. Our decoupled diffusion transformer (DDT) presents a new variant within the diffusion transformer family. It achieves faster convergence by decoupling the low-frequency encoding and the high-frequency decoding. 

\paragraph{Fast Diffusion Training.} To accelerate the training efficiency of diffusion transformers, recent advances have pursued multi-faceted optimizations. Operator-centric approaches~\cite{diffusionrwkv,flowdcn,diffusionssm,dim} leverage efficient attention mechanisms: linear-attention variants~\cite{diffusionrwkv, diffusionssm, dim} reduced quadratic complexity to speed up training, while sparse-attention architectures~\cite{flowdcn} prioritized sparsely relevant token interactions. Resampling approaches~\cite{sd3, minsnr} proposed lognorm sampling~\cite{sd3} or loss reweighting~\cite{minsnr} techniques to stabilize training dynamics. Representation learning enhancement approaches integrate external inductive biases: REPA~\cite{repa}, RCG~\cite{rcg} and DoD~\cite{dod} borrowed vision-specific priors into diffusion training, while masked modeling techniques~\cite{maskdit, mdt} strengthened spatial reasoning by enforcing structured feature completion during denoising. Collectively, these strategies address computational, sampling, and representational bottlenecks.

\section{Preliminary Analysis}
\begin{figure}
    \centering
    \includegraphics[width=1.0\linewidth]{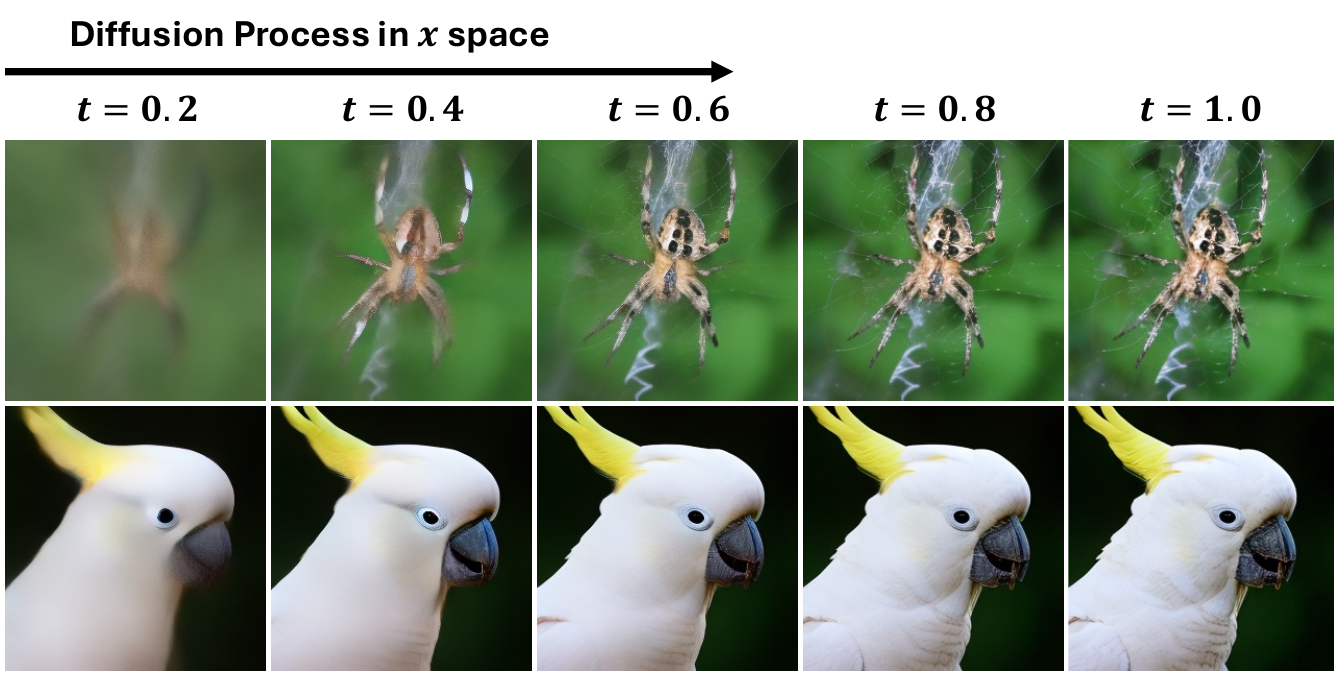}
    \caption{\textbf{The reverse-SDE process~(generation) of SiT-XL/2 in $x$ space.} {\small There is a clear generation process from low frequency to high frequency. Most of the time is spent on generating high-frequency details~(from $t=0.4$ to $t=1.0$). }}
    \label{fig:spec_diffusion}
    \vspace{-1em}
\end{figure}

\begin{figure}
    \centering
    \includegraphics[width=0.49\linewidth]{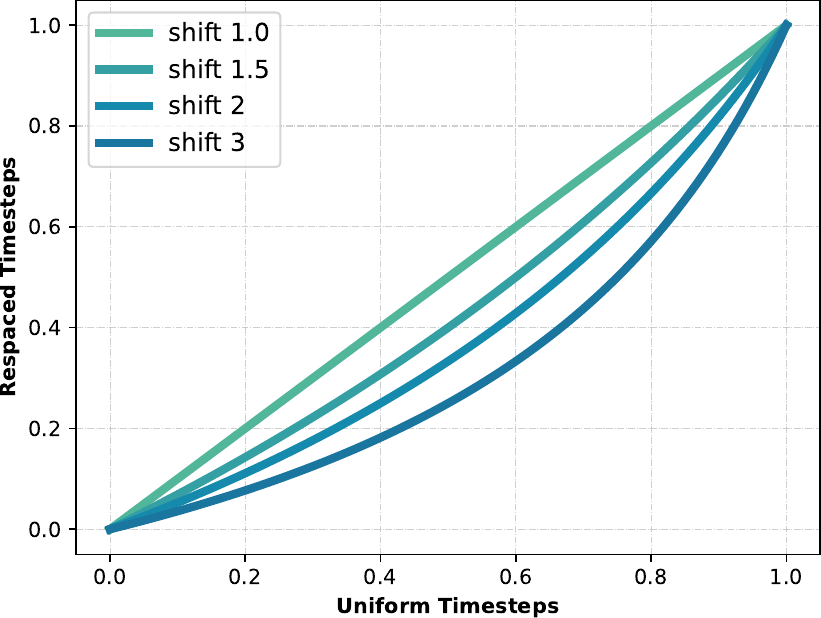}
    \includegraphics[width=0.48\linewidth]{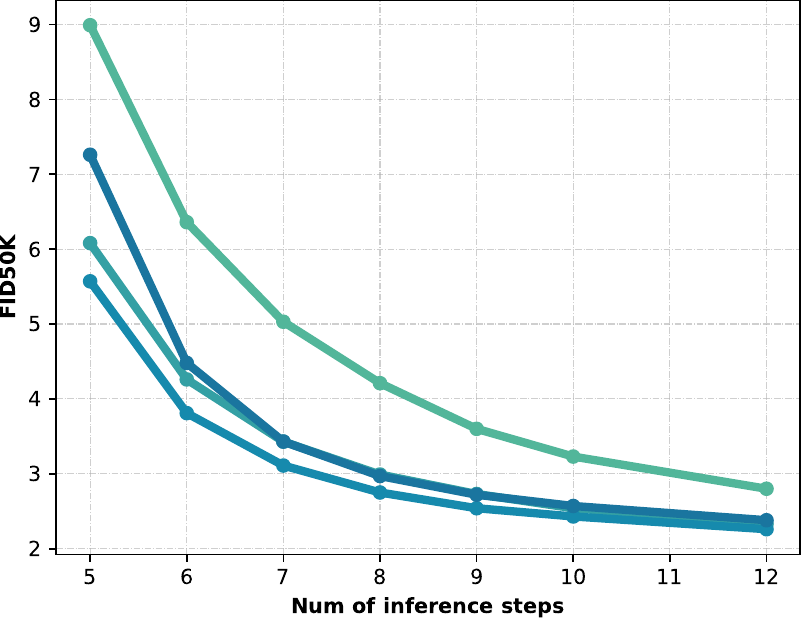}
    \caption{\textbf{The FID50K metric of SiT-XL/2 for different timeshift values.} We employ a $2$-nd order Adams-like solver to collect the performance. Allocating more computation at noisy steps significantly improves the performance.}
    \label{fig:timeshift}
    \vspace{-2em}
\end{figure}

Linear-based flow matching~\cite{flow, flow2, sit} represents a specialized family of diffusion models that we focus on as our primary analytical subject due to its simplicity and efficiency. For the convenience of discussion, in certain situations, diffusion and flow-matching will be used interchangeably. In this framework, $t=0$ corresponds to the pure noise timestep.

As illustrated in \cref{fig:spec_diffusion}, diffusion models perform autoregressive refinement on spectral components~\cite{heat, spec_diffusion}. The diffusion transformer encodes the noisy latent to capture lower-frequency semantics before decoding higher-frequency details. However, this semantics encoding process inevitably attenuates high-frequency information, creating an optimization dilemma. This observation motivates our proposal to decouple the conventional decode-only diffusion transformer into an explicit encoder-decoder architecture.

\begin{lemma}
For a linear flow-matching noise scheduler at timestep $t$, let us denote $K_{freq}$ as the maximum frequency of the clean data ${\bs x}_{data}$. The maximum retained frequency in the noisy latent satisfies:
\begin{equation}
f_{max}(t) > \min\left({\left(\frac{t}{1-t}\right)}^2, K_{freq}\right) .
\end{equation}
\label{lemma:spec_diffusion}
\vspace{-1em}
\end{lemma}

\cref{lemma:spec_diffusion} is directly borrowed from ~\cite{heat, spec_diffusion}, we place the proof of \cref{lemma:spec_diffusion} in Appendix. According to \cref{lemma:spec_diffusion}, as $t$ increases to less noisy timesteps, semantic encoding becomes easier (due to noise reduction) while decoding complexity increases (as residual frequencies grow). Consider the worst-case scenario at denoising step $t$, the diffusion transformer encodes frequencies up to $f_{max}(t)$, to progress to step $s$, it must decode a residual frequency of at least $f_{max}(s) - f_{max}(t)$. Failure to decode these residual frequencies at step $t$ creates a critical bottleneck for progression to subsequent steps. From this perspective, if allocating more of the calculations to more noisy timesteps can lead to an improvement, it means that diffusion transformers struggle with encoding lower frequency to provide semantics. Otherwise, if allocating more of the calculations to less noisy timesteps can lead to an improvement, it means that flow-matching transformers struggle with decoding higher frequency to provide fine details.

To figure out the bottom-necks of current diffusion models, we conducted a targeted experiment using SiT-XL/2 with a second-order Adams-like linear multistep solver. As shown in \cref{fig:timeshift}, by varying the time-shift values, we demonstrate that allocating more computation to early timesteps improves final performance compared to uniform scheduling. This reveals that diffusion models face challenges in more noisy steps. This leads to a key conclusion: \textbf{{Current diffusion transformers are fundamentally constrained by their low-frequency semantic encoding capacity}}. This insight motivates the exploration of encoder-decoder architectures with strategic encoder parameter allocation.

Prior researches further support this perspective. While lightweight diffusion MLP heads demonstrate limited decoding capacity, MAR~\cite{mar} overcomes this limitation through semantic latents produced by its masked backbones, enabling high-quality image generation. Similarly, REPA~\cite{repa} enhances low-frequency encoding through alignment with pre-trained vision foundations~\cite{dinov2}.

\section{Method}
Our decoupled diffusion transformer architecture comprises a condition encoder and a velocity decoder. The condition encoder extracted the low-frequency component from noisy input, class label, and timestep to serve as a self-condition for the velocity decoder; the velocity decoder processed the noisy latent with the self-condition to regress the high-frequency velocity.  We train this model using the established linear flow diffusion framework. For brevity, we designate our model as \textbf{\color{ddt}DDT}~(\textbf{\color{ddt}D}ecoupled \textbf{\color{ddt}D}iffusion \textbf{\color{ddt}T}ransformer).

\subsection{Condition Encoder}
The condition encoder mirrors the architectural design and input structure of DiT/SiT with improved micro-design. It is built with interleaved Attention and FFN blocks, without long residual connections. The encoder processes three inputs, the noisy latent ${\bs x}_t$, timestep $t$, and class label $y$, to extract the self-condition feature ${\bs z}_t$ through a series of stacked Attention and FFN blocks:
\begin{equation}
    {\bs z}_t = \textbf{Encoder}~({\bs x}_t, t, y).
\end{equation}
Specifically, the noisy latent ${\bs x}_t$ are patchfied into continuous tokens and then fed to extract the self-condition ${\bs z}_t$ with aforementioned encoder blocks. The timestep $t$ and class label $y$ serve as external-conditioning information projected into embedding. These external-condition embeddings are progressively injected into the encoded features of ${\bs x}_t$ using AdaLN-Zero\cite{dit} within each encoder block.

To maintain local consistency of ${\bs z}_t$ across adjacent timesteps, we adopt the representation alignment technique from REPA~\cite{repa}. Shown in \cref{eq:encoder_loss}, this method aligns the intermediate feature $\mathbf{h}_i$ from the $i$-th layer in the self-mapping encoder with the DINOv2 representation $r_*$.  Consistent to REPA~\cite{repa}, the $h_{\phi}$ is the learnable projection MLP:
\begin{equation}
    \mathcal{L}_{enc} = 1-\cos(r_*, h_{\phi}(\mathbf{h_i})) .
    \label{eq:encoder_loss}
\end{equation}
This simple regularization accelerates training convergence, as shown in REPA~\cite{repa}, and facilitates local consistency of ${\bs z}_t$ between adjacent steps. It allows sharing the self-condition ${\bs z}_t$ produced by the encoder between adjacent steps. Our experiments demonstrate that this encoder-sharing strategy significantly enhances inference efficiency with only negligible performance degradation.

Additionally, the encoder also receives indirect supervision from the decoder, which we elaborate on later.
\subsection{Velocity Decoder}
The velocity decoder adopts the same architectural design as the condition encoder and consists of several stacked interleaved Attention and FFN blocks, akin to DiT/SiT. It takes the noisy latent ${\bs x}_t$, timestep $t$, and self-conditioning ${\bs z}_t$ as inputs to estimate the velocity ${\bs v}_t$. Unlike the encoder, we assume that class label information is already embedded within ${\bs z}_t$. Thus, only the external-condition timestep $t$ and self-condition feature ${\bs z}_t$ are used as condition inputs for the decoder blocks:
\begin{equation}
    {\bs v}_t = \textbf{Decoder}~({\bs x}_t, t, {\bs z}_t).
\end{equation}
As demonstrated previously, to further improve consistency of self-condition ${\bs z}_t$ between adjacent steps, we employ AdaLN-Zero~\cite{dit} to inject ${\bs z}_t$ into the decoder feature. The decoder is trained with the flow matching loss as shown in \cref{eq:fm_loss}:
\begin{equation}
    \mathcal{L}_{dec} = \mathbb{E} [\int_0^1 ||({\bs x}_{data} - {\epsilon}) - {\bs v}_t||^2 \mathrm{d}t] .
    \label{eq:fm_loss}
\end{equation}

\subsection{Sampling acceleration}
By incorporating explicit representation alignment into the encoder and implicit self-conditioning injection into the decoder, we achieve local consistency of ${\bs z}_t$ across adjacent steps during training~(shown in \cref{fig:zt_sim}). This enables us to share ${\bs z}_t$ within a suitable local range, reducing the computational burden on the self-mapping encoder.

Formally, given total inference steps $N$ and encoder computation bugets $K$, thus the sharing ratio is $1-\frac{K}{N}$, we define $\Phi$ with $|\Phi| = K$ as the set of timesteps where the self-condition is recalculated, as shown in Equation \ref{eq:sharing_encoder}. If the current timestep $t$ is not in $\Phi$, we reuse the previously computed ${\bs z}_{t - \Delta t}$ as ${\bs z}_t$. Otherwise, we recompute ${\bs z}_t$ using the encoder and the current noisy latent ${\bs x}_t$:
\begin{equation}
{\bs z}_{t} = 
\begin{cases}
{\bs z}_{t-\Delta t}, & \text{if } t \notin \Phi \\
 \textbf{Encoder}~({\bs x}_{t}, t, y), & \text{if } t \in \Phi
\end{cases}
\label{eq:sharing_encoder}
\end{equation}
\paragraph{Uniform Encoder Sharing.} This naive approach recaluculate self-condition ${\bs z}_t$ every $\frac{N}{K}$ steps. Previous work, such as DeepCache~\cite{deepcache}, uses this naive handcrafted uniform $\Phi$ set to accelerate UNet models. However, UNet models, trained solely with a denoising loss and lacking robust representation alignment, exhibit less regularized local consistency in deeper features across adjacent steps compared to our DDT model. Also, we will propose a simple and elegant statistic dynamic programming algorithm to construct $\Phi$. Our statistic dynamic programming can exploit the optimal $\Phi$ set optimally compared to the naive approaches~\cite{deepcache}.

\paragraph{Statistic Dynamic Programming.} We construct the statistic similarity matrix of $z_t$ among different steps $\mathbf{S}\in R^{N\times N}$ using cosine distance. The optimal $\Phi$ set would guarantee the total similarity cost $-\sum_{k}^K \sum_{i=\Phi_k}^{\Phi_{k+1}} S[\Phi_k,i]$ achieves global minimal. This question is a well-formed classic minimal sum path problem, it can be solved by dynamic programming. As shown in \cref{eq:sdp}, we donate $\mathbf{C}^k_i$ as cost and $\mathbf{P}^k_i$ as traced path when $\Phi_k = i$.  the state transition function from $\mathbf{C}^{k-1}_j$ to $\mathbf{C}^k_i$ follows:
\begin{align}
    \mathbf{C}^k_i &= \min_{j=0}^i \{\mathbf{C}^{k-1}_j - \Sigma_{l=j}^{i} \mathbf{S}[j, l]\} . \\
    \mathbf{P}^k_i &= \argmin_{j=0}^i \{\mathbf{C}^{k-1}_i - \Sigma_{l=j}^{i} \mathbf{S}[j, l]\} .
    \label{eq:sdp}
\end{align}
After obtaining the cost matrix $\mathbf{C}$ and tracked path $\mathbf{P}$, the optimal $\Phi$ can be solved by backtracking $\mathbf{P}$ from $\mathbf{P}_{N}^K$.

\section{Experiment}
We conduct experiments on 256x256 ImageNet datasets. The total training batch size is set to 256. Consistent with methodological approaches such as SiT~\cite{sit}, DiT~\cite{dit}, and REPA~\cite{repa}, we employed the Adam optimizer with a constant learning rate of 0.0001 throughout the entire training process. To ensure a fair comparative analysis, we did not use gradient clipping and learning rate warm-up techniques. Our default training infrastructure consisted of $16\times$ or $8\times$ A100 GPUs.  For sampling, we take the Euler solver with 250 steps as the default choice. As for the VAE, we take the off-shelf VAE-ft-EMA with a downsample factor of 8 from Huggingface\footnote{\url{https://huggingface.co/stabilityai/sd-vae-ft-ema}}. We report FID~\cite{fid}, sFID~\cite{sfid}, IS~\cite{is}, Precision and Recall~\cite{pr_recall}.
\begin{table*}[t]
\centering
\begin{tabular}{c|c|c|cccc|cccc}
\toprule
& & & \multicolumn{4}{c}{256$\times$256, w/o CFG} & \multicolumn{4}{c}{256$\times$256, w/ CFG}\\
& Params & Epochs
& FID$\downarrow$ & IS$\uparrow$ & Pre.$\uparrow$ & Rec.$\uparrow$
& FID$\downarrow$ & IS$\uparrow$ & Pre.$\uparrow$ & Rec.$\uparrow$ \\
\midrule
\textcolor{lightgray}{MAR-B~\cite{mar}} & \textcolor{lightgray}{208M} & \textcolor{lightgray}{800}
& \textcolor{lightgray}{3.48} & \textcolor{lightgray}{192.4} & \textcolor{lightgray}{0.78} & \textcolor{lightgray}{0.58} 
& \textcolor{lightgray}{2.31} & \textcolor{lightgray}{281.7} & \textcolor{lightgray}{0.82} & \textcolor{lightgray}{0.57} \\
\textcolor{lightgray}{CausalFusion~\cite{causalfusion}} & \textcolor{lightgray}{368M} & \textcolor{lightgray}{800}
& \textcolor{lightgray}{5.12} & \textcolor{lightgray}{166.1} & \textcolor{lightgray}{0.73} & \textcolor{lightgray}{0.66} 
& \textcolor{lightgray}{1.94} & \textcolor{lightgray}{264.4} & \textcolor{lightgray}{0.82} & \textcolor{lightgray}{0.59} \\
LDM-4~\cite{ldm} & 400M & 170
& 10.56 & 103.5 & 0.71 & 0.62 
& 3.6 & 247.7 & {0.87} & 0.48 \\

{\color{ddt} DDT-L~(Ours)} & {\color{ddt} 458M} & {\color{ddt} 80}
& {\color{ddt} 7.98} & {\color{ddt} 128.1} & {\color{ddt} 0.68} & \textbf{\color{ddt} 0.67} 
& \textbf{\color{ddt} 1.64} & \textbf{\color{ddt} 310.5} & {\color{ddt} 0.81} & \textbf{\color{ddt} 0.61} \\

\hline
\textcolor{lightgray}{MAR-L~\cite{mar}} & \textcolor{lightgray}{479M} & \textcolor{lightgray}{800}
& \textcolor{lightgray}{2.6} & \textcolor{lightgray}{221.4} & \textcolor{lightgray}{0.79} & \textcolor{lightgray}{0.60} & 
\textcolor{lightgray}{1.78} & \textcolor{lightgray}{296.0} & \textcolor{lightgray}{0.81} & \textcolor{lightgray}{0.60} \\

\textcolor{lightgray}{VAVAE~\cite{vavae}} & \textcolor{lightgray}{675M} &\textcolor{lightgray}{800}
& \textcolor{lightgray}{2.17} & \textcolor{lightgray}{205.6} & \textcolor{lightgray}{0.77} & \textcolor{lightgray}{0.65} & 
\textcolor{lightgray}{1.35} & \textcolor{lightgray}{295.3} & \textcolor{lightgray}{0.79} & \textcolor{lightgray}{0.65}\\

\textcolor{lightgray}{CausalFusion~\cite{causalfusion}} & \textcolor{lightgray}{676M} & \textcolor{lightgray}{800}
& \textcolor{lightgray}{3.61} & \textcolor{lightgray}{180.9} & \textcolor{lightgray}{0.75} & \textcolor{lightgray}{0.66} 
& \textcolor{lightgray}{1.77} & \textcolor{lightgray}{282.3} & \textcolor{lightgray}{0.82} & \textcolor{lightgray}{0.61} \\

ADM~\cite{adm} & 554M & 400
& 10.94 & - & 0.69 & 0.63 
& 4.59 & 186.7 & 0.82 & 0.52  \\

DiT-XL~\cite{dit} & 675M & 1400
& 9.62 & 121.5 & 0.67 & 0.67 
& 2.27 & 278.2 & {0.83} & 0.57 \\

SiT-XL~\cite{sit} & 675M & 1400
& 8.3 & - & - & - 
& 2.06 & 270.3 & 0.82 & 0.59 \\

ViT-XL~\cite{minsnr} & 451M & 400
& 8.10 & - & - & - 
& 2.06 & - & - & - \\

U-ViT-H/2~\cite{uvit} & 501M & 400
& 6.58 & - & - & - 
& 2.29 & 263.9 & 0.82 & 0.57 \\

MaskDiT~\cite{maskdit} & 675M & 1600
& 5.69 & 178.0 & 0.74 & 0.60 
& 2.28 & 276.6 & 0.80 & 0.61  \\

FlowDCN~\cite{flowdcn} & 618M & 400
& 8.36 & 122.5 & 0.69 & 0.65 
& 2.00 & 263.1 & 0.82 & 0.58 \\

RDM~\cite{rdm} & 553M & /
& 5.27 & 153.4 & 0.75 & 0.62 
& 1.99 & 260.4 & 0.81 & 0.58 \\

REPA~\cite{repa} & 675M & 800
& 5.9  & 157.8 & 0.70 & 0.69 
& 1.42 & 305.7 & 0.80 & 0.64 \\

{\color{ddt} DDT-XL~(Ours)} & {\color{ddt} 675M} & {\color{ddt} 80}
& {\color{ddt} 6.62} & {\color{ddt} 135.2} & {\color{ddt} 0.69} & {\color{ddt} 0.67}
& {\color{ddt} 1.52} & {\color{ddt} 263.7} & {\color{ddt} 0.78} & {\color{ddt} 0.63}\\

{\color{ddt} DDT-XL~(Ours)} & {\color{ddt} 675M} & {\color{ddt} 256}
& {\color{ddt} 6.30} & {\color{ddt} 146.7} & {\color{ddt} 0.68} & {\color{ddt} 0.68} 
& \textbf{\color{ddt} 1.31} & \textbf{\color{ddt} 308.1} & {\color{ddt} 0.78} & {\color{ddt} 0.62} \\

{\color{ddt} DDT-XL~(Ours)} & {\color{ddt} 675M} & {\color{ddt} 400}
 & {\color{ddt} 6.27} & {\color{ddt} 154.7} & {\color{ddt} 0.68} & {\color{ddt} 0.69} 
 & \textbf{\color{ddt} 1.26} & \textbf{\color{ddt} 310.6} & {\color{ddt} 0.79} & \textbf{\color{ddt} 0.65} \\

\bottomrule
\end{tabular}
\vspace{-8pt}
\caption{
\textbf{System performance comparison} on ImageNet $256\times256$ class-conditioned generation. \textcolor{lightgray}{Gray} blocks mean the algorithm uses VAE trained or fine-tuned on ImageNet instead of the off-shelf SD-VAE-f8d4-ft-ema.
\vspace{-8pt}
}
\label{tab:imagenet256_sota}
\end{table*}

\subsection{Improved baselines}
Recent architectural improvements such as SwiGLU~\cite{llama1,llama2}, RoPE~\cite{rope}, and RMSNorm~\cite{llama1, llama2} have been extensively validated in the research community~\cite{visionllama, vavae, fit}. Additionally, lognorm sampling~\cite{sd3} has demonstrated significant benefits for training convergence. Consequently, we developed improved baseline models by incorporating these advanced techniques, drawing inspiration from recent works in the field. The performance of these improved baselines is comprehensively provided in \cref{tab:400k}.
To validate the reliability of our implementation, we also reproduced the results for REPA-B/2, achieving metrics that marginally exceed those originally reported in the REPA\cite{repa}. These reproduction results provide additional confidence in the robustness of our approach.

The improved baselines in our \cref{tab:400k} consistently outperform their predecessors without REPA. However, upon implementing REPA, performance rapidly approaches a saturation point. This is particularly evident in the XL model size, where incremental technique improvements yield diminishingly small gains.
\subsection{Metric comparison with baselines}
We present the performances of different-size models at 400K training steps in \cref{tab:400k}. Our diffusion encoder-decoder transformer(DDT) family demonstrates consistent and significant improvements across various model sizes. Our DDT-B/2(8En4De) model exceeds Improved-REPA-B/2 by 2.8 FID gains. Our DDT-XL/2(22En6De) exceeds REPA-XL/2 by 1.3 FID gains. While the decoder-only diffusion transformers approach performance saturation with REPA\cite{repa}, our DDT models continue to deliver superior results. The incremental technique improvements show diminishing gains, particularly in larger model sizes. However, our DDT models maintain a significant performance advantage, underscoring the effectiveness of our approach.

\begin{table}[ht!]
\centering
\small
\setlength{\tabcolsep}{2pt}
\begin{tabular}{l|ccccc}
\toprule
     Model & FID$\downarrow$ & sFID$\downarrow$ & IS$\uparrow$ & Prec.$\uparrow$ & Rec.$\uparrow$\\
     \midrule
     {\color{gray} SiT-B/2}  \cite{sit} 
     & {\color{gray} 33.0} & {\color{gray} 6.46} & {\color{gray} 43.7} & {\color{gray} 0.53} & {\color{gray} 0.63} \\
     {\color{gray}REPA-B/2} \cite{repa}  
     & {\color{gray} 24.4} & {\color{gray} 6.40} & {\color{gray} 59.9} & {\color{gray} 0.59} & {\color{gray} 0.65} \\
     REPA-B/2(Reproduced)
     & 22.2 & 7.50 & 69.1 & 0.59 & 0.65 \\
{\color{ddt} DDT-B/2$^\dagger$~(8En4De)}
& \textbf{\color{ddt} 21.1} & {\color{ddt} 7.81} & \textbf{\color{ddt} 73.0} & \textbf{\color{ddt} 0.60} &\textbf{\color{ddt}0.65} \\
    \midrule
     Improved-SiT-B/2  
     & 25.1 & 6.54 & 58.8 & 0.57 & 0.64 \\
     Improved-REPA-B/2
     & 19.1 & 6.88 & 76.49 & 0.60 & 0.66 \\
     {\color{ddt} DDT-B/2~(8En4De)}
     & \textbf{\color{ddt} 16.32} & \textbf{\color{ddt} 6.63} & \textbf{\color{ddt} 86.0} & \textbf{\color{ddt} 0.62} & \textbf{\color{ddt} 0.66} \\
     
     \midrule
     {\color{gray} SiT-L/2} \cite{sit} 
     & {\color{gray} 18.8} & {\color{gray} 5.29} & {\color{gray} 72.0} & {\color{gray} 0.64} & {\color{gray} 0.64} \\
     {\color{gray} REPA-L/2} \cite{repa} 
     & {\color{gray}10.0} & {\color{gray} 5.20} & {\color{gray}109.2} & {\color{gray}0.69} & {\color{gray}0.65} \\
     Improved-SiT-L/2
     & 12.7 & 5.48 & 95.7 & 0.65 & 0.65 \\
     Improved-REPA-L/2
     & 9.3 & 5.44 & 116.6 & 0.67 & 0.66 \\
     {\color{ddt} DDT-L/2~(20En4De)}
     & \textbf{\color{ddt} 7.98} & {\color{ddt} 5.50} & \textbf{\color{ddt} 128.1} & \textbf{\color{ddt} 0.68} & \textbf{\color{ddt} 0.67} \\
     
     \midrule
     {\color{gray} SiT-XL/2} \cite{sit} 
     & {\color{gray}17.2} & {\color{gray}5.07} & {\color{gray}76.52} & {\color{gray}0.65} & {\color{gray}0.63}\\
     {\color{gray}REPA-XL/2} \cite{repa} 
     & {\color{gray}7.9} & {\color{gray}5.06} & {\color{gray}122.6} & {\color{gray}0.70} & {\color{gray}0.65}\\
      Improved-SiT-XL/2
     & 10.9 & 5.3 & 103.4 & 0.66 & 0.65\\
     Improved-REPA-XL/2
     & 8.14 & 5.34 & 124.9 & 0.68 & 0.67\\
     {\color{ddt} DDT-XL/2~(22En6De)}
     & \textbf{\color{ddt} 6.62} & \textbf{\color{ddt} 4.86} & \textbf{\color{ddt} 135.1} & \textbf{\color{ddt} 0.69} & \textbf{\color{ddt} 0.67} \\
\bottomrule
\end{tabular}
\caption{\textbf{Metrics of $400K$ training steps with different model sizes.} {\small All results are reported without classifier-free guidance. {\color{gray} gray} means metrics are copied from the original paper, otherwise it is produced by our codebase. By default, our DDT models are built on improved baselines. DDT$^\dagger$ means model built on naive baseline without architecture improvement and lognorm sampling, consistent to REPA.  Our DDT models consistently outperformed their counterparts. }}
\label{tab:400k}
\vspace{-1em}
\end{table}

\subsection{System level comparision}
\paragraph{ImageNet $256\times256$.} We report the final metrics of DDT-XL/2 (22En6De) and DDT-L/2 (20En4De) at \cref{tab:imagenet256_sota}. Our DDT models demonstrate exceptional efficiency, achieving convergence in approximately $\frac{1}{4}$ of the total epochs compared to REPA~\cite{repa} and other diffusion transformer models. In order to maintain methodological consistency with REPA, we employed the classifier-free guidance with 2.0 in the interval $[0.3, 1]$, Our models delivered impressive results: DDT-L/2 achieved 1.64 FID, and DDT-XL/2 got 1.52 FID within just 80 epochs. By extending training to 256 epochs—still significantly more efficient than traditional 800-epoch approaches—our DDT-XL/2 established a new state-of-the-art benchmark of 1.31 FID on ImageNet 256×256, decisively outperforming previous diffusion transformer methodologies. To extend training to $400$ epochs, our DDT-XL/2(22En6De) achieves 1.26 FID, nearly reaching the upper limit of SD-VAE-ft-EMA-f8d4, which has a 1.20 rFID on ImageNet$256\time256$.  
\paragraph{ImageNet $512\times512$} We provide the final metrics of DDT-XL/2 at \cref{tab:imagenet512_sota}. To validate the superiority of our DDT model, we take our DDT-XL/2 trained on ImageNet~$256\times256$ under 256 epochs as the initialization, fine-tune out DDT-XL/2 on ImageNet~$512\times512$ for $100K$ steps. We adopt the aforementioned interval guidance~\cite{interval_guidance} and we achieved a remarkable state-of-the-art performance of 1.90 FID, decisively outperforming REPA by a significant 0.28 performance margin. In \cref{tab:imagenet512_sota}, some metrics exhibit subtle degradation, we attribute this to potentially insufficient fine-tuning. When allocating more training iterations to DDT-XL/2, it achieves $1.28$ FID at 500K steps with CFG3.0 within the time interval $[0.3, 1.0]$. 
\begin{table}[t] 
\centering
\small 
\setlength{\tabcolsep}{3.5pt}
\begin{tabular}{l|ccccc}
\toprule
& \multicolumn{5}{c}{\bf{ImageNet} $512\times512$} \\
\toprule
Model & FID$\downarrow$  & sFID$\downarrow$  & IS$\uparrow$  & Pre.$\uparrow$ & Rec.$\uparrow$ \\
\midrule
BigGAN-deep~\cite{largegan} & 8.43 & 8.13 & 177.90 & 0.88 & 0.29 \\
StyleGAN-XL~\cite{styleganxl} & 2.41 & 4.06 & 267.75 & 0.77 & 0.52 \\
\midrule
ADM-G~\cite{adm} & 7.72 & 6.57 & 172.71 & {0.87} &
0.42 \\
ADM-G, ADM-U & 3.85 & 5.86 & 221.72 & 0.84 & 0.53\\
{DiT-XL/2}~\cite{dit} & {3.04} & {5.02} & {240.82} & 0.84 & 0.54 \\
{SiT-XL/2} ~\cite{sit}  & 2.62 &  4.18 & 252.21 & 0.84 & 0.57 \\
{REPA-XL/2} ~\cite{repa}  & 2.08 & 4.19 & 274.6 & 0.83 & 0.58 \\
{FlowDCN-XL/2} ~\cite{flowdcn} & {2.44} & {4.53} & {252.8} & 0.84 & 0.54 \\





{\color{ddt} DDT-XL/2~(500K)} & \textbf{\color{ddt} 1.28} & {\color{ddt} 4.22} & \textbf{\color{ddt}305.1} & {\color{ddt} 0.80} & \textbf{\color{ddt} 0.63} \\

\bottomrule
\end{tabular}
\caption{\textbf{Benchmarking class-conditional image generation on ImageNet 512$\times$512.} {\small Our DDT-XL/2($512\times512$) is fine-tuned from the same model trained on $256\times256$ resolution setting of 1.28M steps. We adopt the interval guidance with interval $[0.3, 1]$ and CFG of 3.0}}
\label{tab:imagenet512_sota}
\vspace{-2em}
\end{table}

\subsection{Acceleration by Encoder sharing}
\begin{figure}
    \centering
    \includegraphics[width=\linewidth]{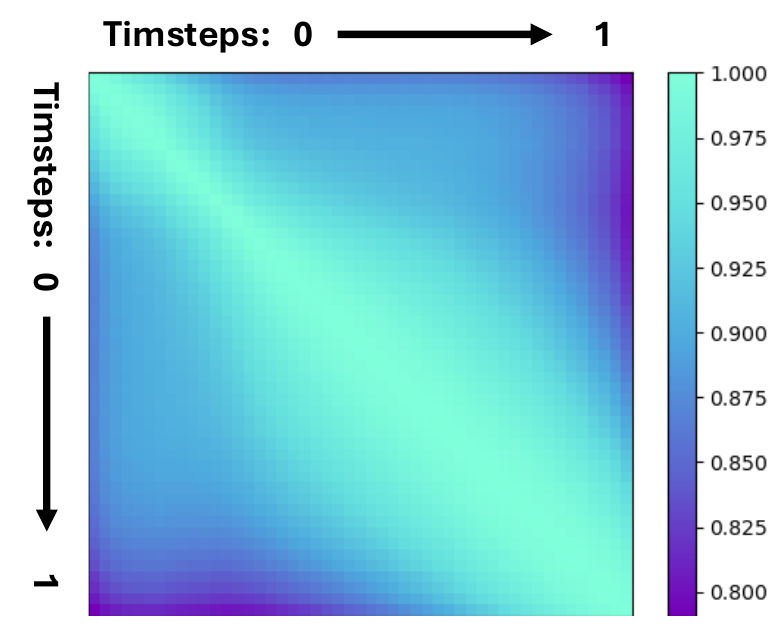}
    \caption{\textbf{The cosine similarity of self-condition feature $\bs z_t$ from encoder between different timesteps.}~{\small There is a strong correlation between adjacent steps, indicating the redundancy.}}
    \label{fig:zt_sim}
    \vspace{-1em}
\end{figure}

As illustrated in \cref{fig:zt_sim}, there is a strong local consistency of the self-condition in our condition encoder. Even ${\bs z}_{t=0}$ has a strong similarity above 0.8 with ${\bs z}_{t=1}$. This consistency provides an opportunity to speed up inference by sharing the encoder between adjacent steps.

\begin{figure}
    \centering
    \includegraphics[width=\linewidth]{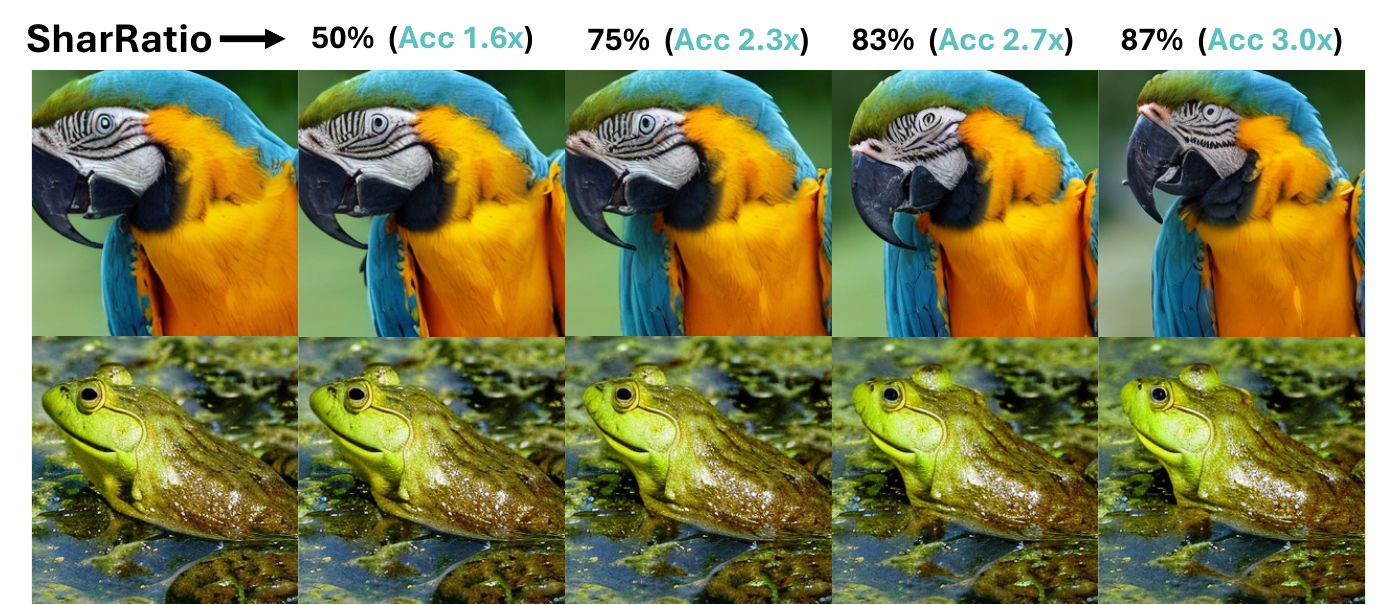}
    \caption{\textbf{Sharing the self-condition $z_t$ in adjacent steps significant speedup the inference.}{\small We tried various sharing frequency configurations. There is marginal visual quality down-gradation when the sharing frequency is reasonable.}}
    \label{fig:sharing_encoder}
    \vspace{-1.5em}
\end{figure}

We employed the simple uniform encoder sharing strategy and the new novel statistics dynamic programming strategy. Specifically, for the uniform strategy, we only recalculate the self-condition ${\bs z}_t$ every $K$ steps. For statistics dynamic programming, we solve the aforementioned minimal sum path on the similarity matrix by dynamic programming and recalculate ${\bs z}_t$ according to the solved strategy.
As shown in \cref{fig:sharing_encoder}, there is a significant inference speedup nearly without visual quality loss when $K$ is smaller than 6. As shown in \cref{tab:sharing_encoder}, the metrics loss is still marginal, while the inference speedup is significant. The novel statistics dynamic programming slightly outperformed the naive uniform strategy with less FID drop.
\begin{table}[ht!]
\centering
\small
\setlength{\tabcolsep}{2pt}
\begin{tabular}{lc|c|cccccccc}
\toprule
SharRatio & Acc & $\Phi$ & FID$\downarrow$  & sFID$\downarrow$ & IS$\uparrow$ & Prec.$\uparrow$ & Rec.$\uparrow$\\
\midrule
0.00 & $1.0\times$ & Uniform & 1.31 & 4.62 & 308.1 & 0.78  & 0.66 \\
0.50 & $1.6\times$ & Uniform & 1.31 & 4.48 & 300.5 & 0.78 & 0.65 \\
0.66 & $1.9\times$ & Uniform & 1.32 & 4.46 & 301.2 & 0.78 & 0.65 \\
0.75 & $2.3\times$ & Uniform & 1.34 & 4.43 & 302.7 & 0.78 & 0.65 \\ 
\midrule
\multirow{2}{*}{0.80} & \multirow{2}{*}{$2.6\times$} & 
Uniform & 1.36 & 4.40 & 303.3 & 0.78 & 0.64 \\ 
& & 
StatisticDP & 1.33 & 4.37 & 301.7 & 0.78 & 0.64 \\ 
\midrule
\multirow{2}{*}{0.83} & \multirow{2}{*}{$2.7\times$} & 
Uniform & 1.37 & 4.41 & 302.8 & 0.78 & 0.64 \\ 
&  & 
StatisticDP & 1.36 & 4.35 & 300.3 & 0.78 & 0.64 \\ 
\midrule
\multirow{2}{*}{0.87} & \multirow{2}{*}{$3.0\times$} & Uniform & 1.42 & 4.43 & 302.8 & 0.78 & 0.64 \\ 
&  & StatisticDP & 1.40 & 4.35 & 302.4 & 0.78 & 0.64 \\ 
\bottomrule
\end{tabular}
\caption{\textbf{Metrics of $400K$ training steps with different model sizes.} {\small All results are reported without classifier-free guidance. {\color{gray} gray} means metrics are copied from the original paper, otherwise it is produced by our codebase. Our DDT models consistently outperformed its counterparts }}
\label{tab:sharing_encoder}
\vspace{-1em}
\end{table}

\subsection{Ablations}
\begin{figure*}
    \centering
    \subfloat{
    \includegraphics[width=0.32\linewidth]{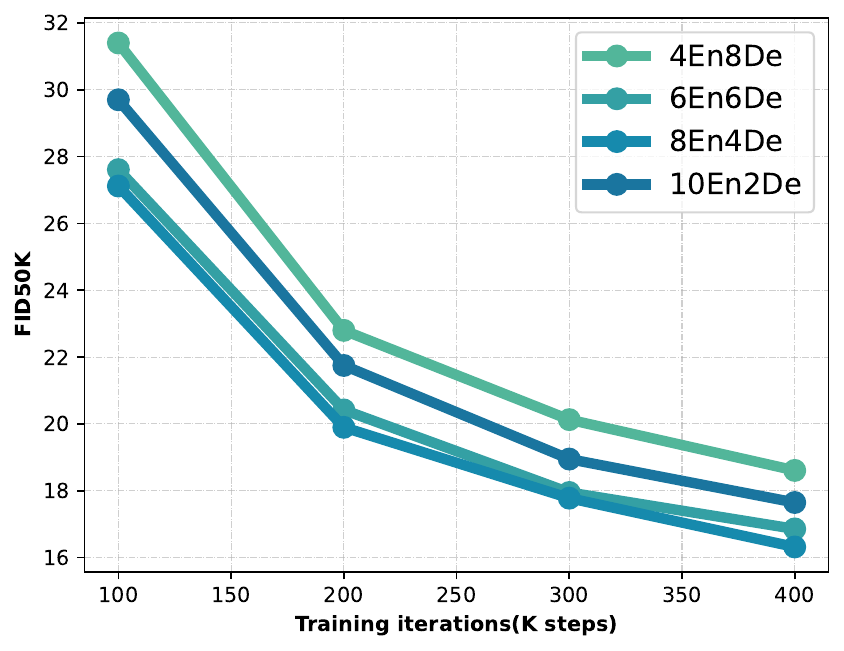}
    }
    \subfloat{
    \includegraphics[width=0.32\linewidth]{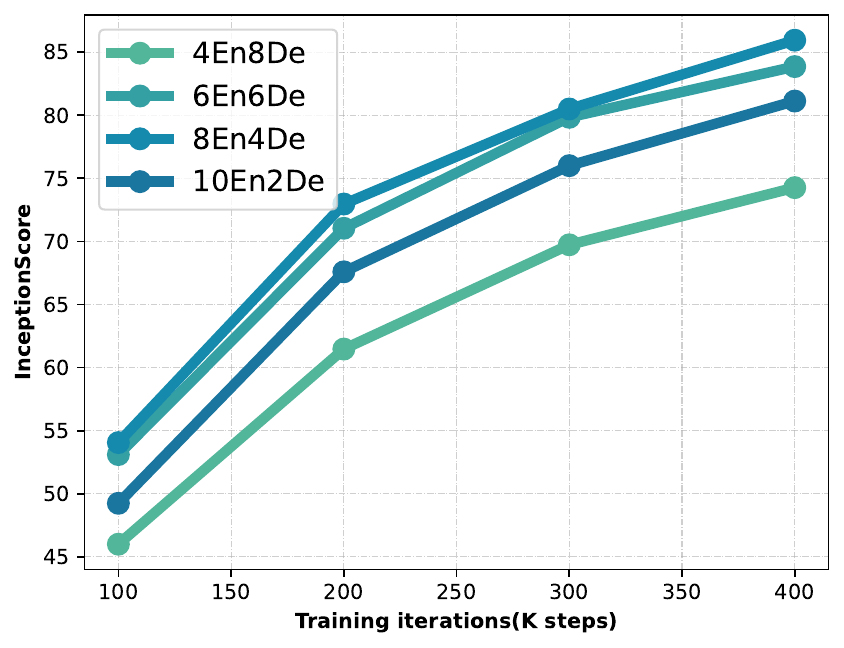}
    }
    \subfloat{
    \includegraphics[width=0.32\linewidth]{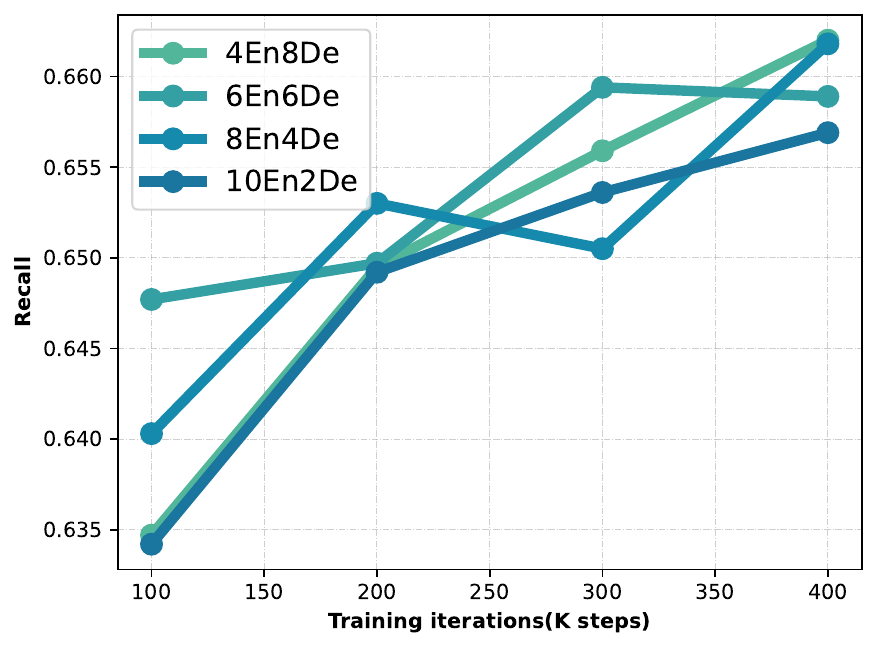}
    }
    \caption{\textbf{The DDT-B/2 built upon Improved-baselines under various Encoder and Decoder layer ratio.}~{\small DDT-B/2(8En4De) achieves much faster convergence speed and better performance.}}
    \label{tab:layer_ratio_base++}
    \vspace{-1.5em}
\end{figure*}
\begin{figure*}
    \centering
    \subfloat{
    \includegraphics[width=0.32\linewidth]{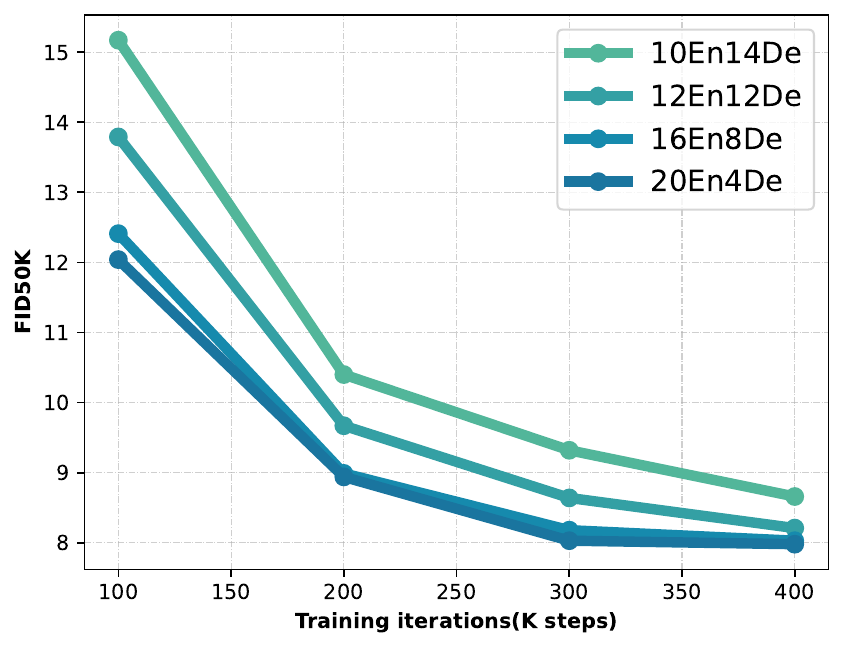}
    }
    \subfloat{
    \includegraphics[width=0.32\linewidth]{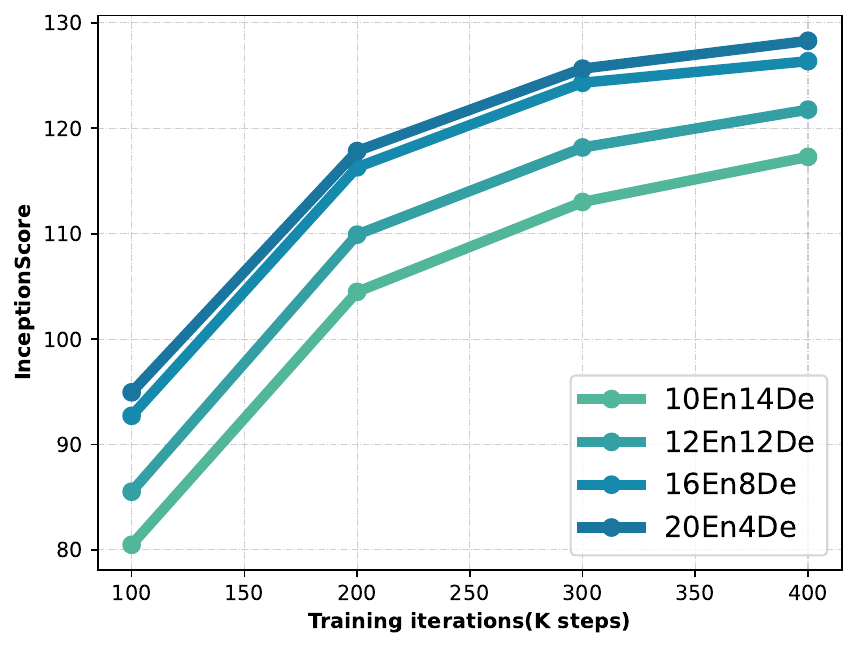}
    }
    \subfloat{
    \includegraphics[width=0.32\linewidth]{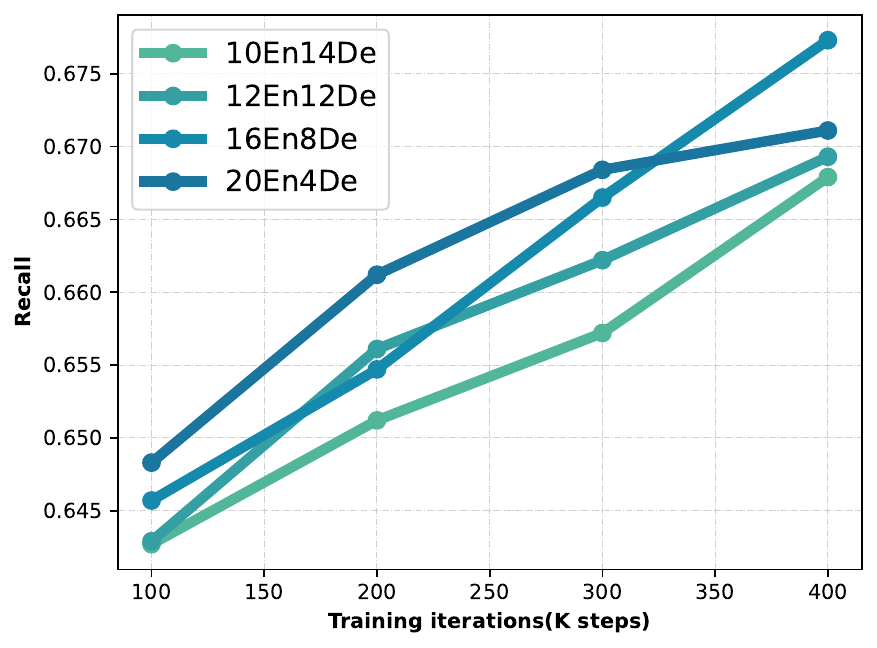}
    }
    \caption{\textbf{The DDT-L/2 built upon Improved-baselines under various Encoder and Decoder layer ratio.}~{\small DDT-L/2 prefers an unexpected aggressive encoder-deocder ratio DDT-L/2(20En4De) achieves much faster convergence speed and better performance.}}
    \label{tab:layer_ratio_large++}
    \vspace{-1.5em}
\end{figure*}
We conduct ablation studies on ImageNet $256\times256$ with DDT-B/2 and DDT-L/2. For sampling, we take the Euler solver with 250 steps as the default choice without classifier-free guidance. For training, we train each model with 80 epochs(400k steps), and the batch size is set to 256.
\paragraph{Encoder-Decoder Ratio} we systematically explored ratios ranging from $2:1$ to $5:1$ across different model sizes. in \cref{tab:layer_ratio_base++} and \cref{tab:layer_ratio_large++}. Our notation {\color{ddt}$m$}En{\color{ddt}$n$}De represents models with $m$ encoder layers and $n$ decoder layers. The investigation experiments in \cref{tab:layer_ratio_base++} and \cref{tab:layer_ratio_large++} revealed critical insights into architectural optimization. We observed that \textbf{a larger encoder is beneficial for further improving the performance as the model size increases}. For the Base model in \cref{tab:layer_ratio_base++}, the optimal configuration emerged as 8 encoder layers and 4 decoder layers, delivering superior performance and convergence speed. Notably, the Large model in \cref{tab:layer_ratio_large++} exhibited a distinct preference, achieving peak performance with 20 encoder layers and 4 decoder layers, an unexpectedly aggressive encoder-decoder ratio. This unexpected discovery motivates us to scale the layer ratio in DDT-XL/2 to 22 encoder layers and 6 decoders to explore the performance upper limits of diffusion transformers.

\paragraph{Decoder Block types.} In our investigation of decoder block types and their impact on high-frequency decoding performance, we systematically evaluated multiple architectural configurations. Our comprehensive assessment included alternative approaches such as simple 3×3 convolution blocks and naive MLP blocks.  As shown in \cref{tab:decoder_block}, the default~(Attention with the MLP) setting achieves better results. Thanks to the encoder-decoder design, naive Conv blocks even achieve comparable results.

\begin{table}
\centering
\small
\setlength{\tabcolsep}{3pt}
\begin{tabular}{l|ccccc}
\toprule
     DecoderBlock & FID$\downarrow$ & sFID$\downarrow$ & IS$\uparrow$ & Prec.$\uparrow$ & Rec.$\uparrow$\\
     \midrule
Conv+MLP & 16.96 & 7.33 & 85.1 & 0.62 & 0.65 \\
MLP+MLP & 24.13 & 7.89 & 65.0 & 0.57 & 0.65 \\
{\color{ddt} Attn+MLP} & {\color{ddt}16.32} & {\color{ddt}6.63} & {\color{ddt}86.0} & {\color{ddt}0.62} & {\color{ddt}0.66} \\
\bottomrule
\end{tabular}
\caption{\textbf{Metrics of $400K$ training steps on DDT-B/2(8En4De) with different decoder blocks.} {\small All results are reported without classifier-free guidance. The Default Attention + MLP configuration achieves best performance.}}
\label{tab:decoder_block}
\vspace{-1em}
\end{table}
\section{Conclusion}
In this paper, we have introduced a novel Decoupled Diffusion Transformer, which rethinks the optimization dilemma of the traditional diffusion transformer. By decoupling the low-frequency encoding and high-frequency decoding into dedicated components, we effectively resolved the optimization dilemma that has constrained diffusion transformer. Furthermore, we discovered that increasing the encoder capacity relative to the decoder yields increasingly beneficial results as the overall model scale grows. This insight provides valuable guidance for future model scaling efforts. Our experiments demonstrate that our DDT-XL/2 (22En6De) with an unexpected aggressive encoder-decoder layer ratio achieves great performance while requiring only 256 training epochs. This significant improvement in efficiency addresses one of the primary limitations of diffusion models: their lengthy training requirements. The decoupled architecture also presents opportunities for inference optimization through our proposed encoder result sharing mechanism. Our statistical dynamic programming approach for determining optimal sharing strategies enables faster inference while minimizing quality degradation, demonstrating that architectural innovations can yield benefits beyond their primary design objectives.

{
    \small
    \bibliographystyle{ieeenat_fullname}
    \bibliography{main}

\begin{thebibliography}{54}
\providecommand{\natexlab}[1]{#1}
\providecommand{\url}[1]{\texttt{#1}}
\expandafter\ifx\csname urlstyle\endcsname\relax
  \providecommand{\doi}[1]{doi: #1}\else
  \providecommand{\doi}{doi: \begingroup \urlstyle{rm}\Url}\fi

\bibitem[Agarwal et~al.(2025)Agarwal, Ali, Bala, Balaji, Barker, Cai, Chattopadhyay, Chen, Cui, Ding, et~al.]{cosmos}
Niket Agarwal, Arslan Ali, Maciej Bala, Yogesh Balaji, Erik Barker, Tiffany Cai, Prithvijit Chattopadhyay, Yongxin Chen, Yin Cui, Yifan Ding, et~al.
\newblock Cosmos world foundation model platform for physical ai.
\newblock \emph{arXiv preprint arXiv:2501.03575}, 2025.

\bibitem[Bao et~al.(2023)Bao, Nie, Xue, Cao, Li, Su, and Zhu]{uvit}
Fan Bao, Shen Nie, Kaiwen Xue, Yue Cao, Chongxuan Li, Hang Su, and Jun Zhu.
\newblock All are worth words: A vit backbone for diffusion models.
\newblock In \emph{Proceedings of the IEEE/CVF Conference on Computer Vision and Pattern Recognition}, pages 22669--22679, 2023.

\bibitem[Brock et~al.(2018)Brock, Donahue, and Simonyan]{largegan}
Andrew Brock, Jeff Donahue, and Karen Simonyan.
\newblock Large scale gan training for high fidelity natural image synthesis.
\newblock \emph{arXiv preprint arXiv:1809.11096}, 2018.

\bibitem[Carion et~al.(2020)Carion, Massa, Synnaeve, Usunier, Kirillov, and Zagoruyko]{detr}
Nicolas Carion, Francisco Massa, Gabriel Synnaeve, Nicolas Usunier, Alexander Kirillov, and Sergey Zagoruyko.
\newblock End-to-end object detection with transformers.
\newblock In \emph{European conference on computer vision}, pages 213--229. Springer, 2020.

\bibitem[Chang et~al.(2022)Chang, Zhang, Jiang, Liu, and Freeman]{maskgit}
Huiwen Chang, Han Zhang, Lu Jiang, Ce Liu, and William~T Freeman.
\newblock Maskgit: Masked generative image transformer.
\newblock In \emph{Proceedings of the IEEE/CVF Conference on Computer Vision and Pattern Recognition}, pages 11315--11325, 2022.

\bibitem[Chen et~al.(2023)Chen, Yu, Ge, Yao, Xie, Wu, Wang, Kwok, Luo, Lu, et~al.]{pixart}
Junsong Chen, Jincheng Yu, Chongjian Ge, Lewei Yao, Enze Xie, Yue Wu, Zhongdao Wang, James Kwok, Ping Luo, Huchuan Lu, et~al.
\newblock Pixart-$\backslash$alpha: Fast training of diffusion transformer for photorealistic text-to-image synthesis.
\newblock \emph{arXiv preprint arXiv:2310.00426}, 2023.

\bibitem[Chen et~al.(2024)Chen, Ge, Xie, Wu, Yao, Ren, Wang, Luo, Lu, and Li]{pixart_sigma}
Junsong Chen, Chongjian Ge, Enze Xie, Yue Wu, Lewei Yao, Xiaozhe Ren, Zhongdao Wang, Ping Luo, Huchuan Lu, and Zhenguo Li.
\newblock Pixart-$\backslash$sigma: Weak-to-strong training of diffusion transformer for 4k text-to-image generation.
\newblock \emph{arXiv preprint arXiv:2403.04692}, 2024.

\bibitem[Chu et~al.(2024)Chu, Su, Zhang, and Shen]{visionllama}
Xiangxiang Chu, Jianlin Su, Bo Zhang, and Chunhua Shen.
\newblock Visionllama: A unified llama interface for vision tasks.
\newblock \emph{arXiv preprint arXiv:2403.00522}, 2024.

\bibitem[Deng et~al.(2024)Deng, Zh, Li, Guan, and Fan]{causalfusion}
Chaorui Deng, Deyao Zh, Kunchang Li, Shi Guan, and Haoqi Fan.
\newblock Causal diffusion transformers for generative modeling.
\newblock \emph{arXiv preprint arXiv:2412.12095}, 2024.

\bibitem[Dhariwal and Nichol(2021)]{adm}
Prafulla Dhariwal and Alexander Nichol.
\newblock Diffusion models beat gans on image synthesis.
\newblock \emph{Advances in neural information processing systems}, 34:\penalty0 8780--8794, 2021.

\bibitem[Dieleman(2024)]{spec_diffusion}
Sander Dieleman.
\newblock Diffusion is spectral autoregression, 2024.

\bibitem[Esser et~al.(2024)Esser, Kulal, Blattmann, Entezari, M{\"u}ller, Saini, Levi, Lorenz, Sauer, Boesel, et~al.]{sd3}
Patrick Esser, Sumith Kulal, Andreas Blattmann, Rahim Entezari, Jonas M{\"u}ller, Harry Saini, Yam Levi, Dominik Lorenz, Axel Sauer, Frederic Boesel, et~al.
\newblock Scaling rectified flow transformers for high-resolution image synthesis.
\newblock \emph{arXiv preprint arXiv:2403.03206}, 2024.

\bibitem[Fei et~al.(2024)Fei, Fan, Yu, Li, and Huang]{diffusionrwkv}
Zhengcong Fei, Mingyuan Fan, Changqian Yu, Debang Li, and Junshi Huang.
\newblock Diffusion-rwkv: Scaling rwkv-like architectures for diffusion models.
\newblock \emph{arXiv preprint arXiv:2404.04478}, 2024.

\bibitem[Gao et~al.(2023{\natexlab{a}})Gao, Zhou, Cheng, and Yan]{maskdit}
Shanghua Gao, Pan Zhou, Ming-Ming Cheng, and Shuicheng Yan.
\newblock Masked diffusion transformer is a strong image synthesizer.
\newblock In \emph{Proceedings of the IEEE/CVF international conference on computer vision}, pages 23164--23173, 2023{\natexlab{a}}.

\bibitem[Gao et~al.(2023{\natexlab{b}})Gao, Zhou, Cheng, and Yan]{mdt}
Shanghua Gao, Pan Zhou, Ming-Ming Cheng, and Shuicheng Yan.
\newblock Masked diffusion transformer is a strong image synthesizer.
\newblock In \emph{Proceedings of the IEEE/CVF International Conference on Computer Vision}, pages 23164--23173, 2023{\natexlab{b}}.

\bibitem[Hang et~al.(2023)Hang, Gu, Li, Bao, Chen, Hu, Geng, and Guo]{minsnr}
Tiankai Hang, Shuyang Gu, Chen Li, Jianmin Bao, Dong Chen, Han Hu, Xin Geng, and Baining Guo.
\newblock Efficient diffusion training via min-snr weighting strategy.
\newblock In \emph{Proceedings of the IEEE/CVF international conference on computer vision}, pages 7441--7451, 2023.

\bibitem[He et~al.(2022)He, Chen, Xie, Li, Doll{\'a}r, and Girshick]{mae}
Kaiming He, Xinlei Chen, Saining Xie, Yanghao Li, Piotr Doll{\'a}r, and Ross Girshick.
\newblock Masked autoencoders are scalable vision learners.
\newblock In \emph{Proceedings of the IEEE/CVF conference on computer vision and pattern recognition}, pages 16000--16009, 2022.

\bibitem[Heusel et~al.(2017)Heusel, Ramsauer, Unterthiner, Nessler, and Hochreiter]{fid}
Martin Heusel, Hubert Ramsauer, Thomas Unterthiner, Bernhard Nessler, and Sepp Hochreiter.
\newblock Gans trained by a two time-scale update rule converge to a local nash equilibrium.
\newblock \emph{Advances in neural information processing systems}, 30, 2017.

\bibitem[Ho et~al.(2020)Ho, Jain, and Abbeel]{ddpm}
Jonathan Ho, Ajay Jain, and Pieter Abbeel.
\newblock Denoising diffusion probabilistic models.
\newblock \emph{Advances in neural information processing systems}, 33:\penalty0 6840--6851, 2020.

\bibitem[Hong et~al.(2022)Hong, Ding, Zheng, Liu, and Tang]{cogvideo}
Wenyi Hong, Ming Ding, Wendi Zheng, Xinghan Liu, and Jie Tang.
\newblock Cogvideo: Large-scale pretraining for text-to-video generation via transformers.
\newblock \emph{arXiv preprint arXiv:2205.15868}, 2022.

\bibitem[Karras et~al.(2022)Karras, Aittala, Aila, and Laine]{edm}
Tero Karras, Miika Aittala, Timo Aila, and Samuli Laine.
\newblock Elucidating the design space of diffusion-based generative models.
\newblock \emph{Advances in Neural Information Processing Systems}, 35:\penalty0 26565--26577, 2022.

\bibitem[Kingma and Ba(2014)]{adam}
Diederik~P Kingma and Jimmy Ba.
\newblock Adam: A method for stochastic optimization.
\newblock \emph{arXiv preprint arXiv:1412.6980}, 2014.

\bibitem[Kirillov et~al.(2023)Kirillov, Mintun, Ravi, Mao, Rolland, Gustafson, Xiao, Whitehead, Berg, Lo, et~al.]{sam}
Alexander Kirillov, Eric Mintun, Nikhila Ravi, Hanzi Mao, Chloe Rolland, Laura Gustafson, Tete Xiao, Spencer Whitehead, Alexander~C Berg, Wan-Yen Lo, et~al.
\newblock Segment anything.
\newblock In \emph{Proceedings of the IEEE/CVF international conference on computer vision}, pages 4015--4026, 2023.

\bibitem[Kong et~al.(2024)Kong, Tian, Zhang, Min, Dai, Zhou, Xiong, Li, Wu, Zhang, et~al.]{hunyuanvideo}
Weijie Kong, Qi Tian, Zijian Zhang, Rox Min, Zuozhuo Dai, Jin Zhou, Jiangfeng Xiong, Xin Li, Bo Wu, Jianwei Zhang, et~al.
\newblock Hunyuanvideo: A systematic framework for large video generative models.
\newblock \emph{arXiv preprint arXiv:2412.03603}, 2024.

\bibitem[Kynk{\"a}{\"a}nniemi et~al.(2019)Kynk{\"a}{\"a}nniemi, Karras, Laine, Lehtinen, and Aila]{pr_recall}
Tuomas Kynk{\"a}{\"a}nniemi, Tero Karras, Samuli Laine, Jaakko Lehtinen, and Timo Aila.
\newblock Improved precision and recall metric for assessing generative models.
\newblock \emph{Advances in neural information processing systems}, 32, 2019.

\bibitem[Kynk{\"a}{\"a}nniemi et~al.(2024)Kynk{\"a}{\"a}nniemi, Aittala, Karras, Laine, Aila, and Lehtinen]{interval_guidance}
Tuomas Kynk{\"a}{\"a}nniemi, Miika Aittala, Tero Karras, Samuli Laine, Timo Aila, and Jaakko Lehtinen.
\newblock Applying guidance in a limited interval improves sample and distribution quality in diffusion models.
\newblock \emph{arXiv preprint arXiv:2404.07724}, 2024.

\bibitem[Li et~al.(2024)Li, Katabi, and He]{rcg}
Tianhong Li, Dina Katabi, and Kaiming He.
\newblock Return of unconditional generation: A self-supervised representation generation method.
\newblock \emph{Advances in Neural Information Processing Systems}, 37:\penalty0 125441--125468, 2024.

\bibitem[Li et~al.(2025)Li, Tian, Li, Deng, and He]{mar}
Tianhong Li, Yonglong Tian, He Li, Mingyang Deng, and Kaiming He.
\newblock Autoregressive image generation without vector quantization.
\newblock \emph{Advances in Neural Information Processing Systems}, 37:\penalty0 56424--56445, 2025.

\bibitem[Lipman et~al.(2022)Lipman, Chen, Ben-Hamu, Nickel, and Le]{flow2}
Yaron Lipman, Ricky~TQ Chen, Heli Ben-Hamu, Maximilian Nickel, and Matt Le.
\newblock Flow matching for generative modeling.
\newblock \emph{arXiv preprint arXiv:2210.02747}, 2022.

\bibitem[Liu et~al.(2022)Liu, Gong, and Liu]{flow}
Xingchao Liu, Chengyue Gong, and Qiang Liu.
\newblock Flow straight and fast: Learning to generate and transfer data with rectified flow.
\newblock \emph{arXiv preprint arXiv:2209.03003}, 2022.

\bibitem[Lu et~al.(2024)Lu, Wang, Huang, Wu, Liu, Ouyang, and Bai]{fit}
Zeyu Lu, Zidong Wang, Di Huang, Chengyue Wu, Xihui Liu, Wanli Ouyang, and Lei Bai.
\newblock Fit: Flexible vision transformer for diffusion model.
\newblock \emph{arXiv preprint arXiv:2402.12376}, 2024.

\bibitem[Ma et~al.(2024{\natexlab{a}})Ma, Goldstein, Albergo, Boffi, Vanden-Eijnden, and Xie]{sit}
Nanye Ma, Mark Goldstein, Michael~S Albergo, Nicholas~M Boffi, Eric Vanden-Eijnden, and Saining Xie.
\newblock Sit: Exploring flow and diffusion-based generative models with scalable interpolant transformers.
\newblock \emph{arXiv preprint arXiv:2401.08740}, 2024{\natexlab{a}}.

\bibitem[Ma et~al.(2024{\natexlab{b}})Ma, Fang, and Wang]{deepcache}
Xinyin Ma, Gongfan Fang, and Xinchao Wang.
\newblock Deepcache: Accelerating diffusion models for free.
\newblock In \emph{Proceedings of the IEEE/CVF conference on computer vision and pattern recognition}, pages 15762--15772, 2024{\natexlab{b}}.

\bibitem[Nash et~al.(2021)Nash, Menick, Dieleman, and Battaglia]{sfid}
Charlie Nash, Jacob Menick, Sander Dieleman, and Peter~W Battaglia.
\newblock Generating images with sparse representations.
\newblock \emph{arXiv preprint arXiv:2103.03841}, 2021.

\bibitem[Oquab et~al.(2023)Oquab, Darcet, Moutakanni, Vo, Szafraniec, Khalidov, Fernandez, Haziza, Massa, El-Nouby, et~al.]{dinov2}
Maxime Oquab, Timoth{\'e}e Darcet, Th{\'e}o Moutakanni, Huy Vo, Marc Szafraniec, Vasil Khalidov, Pierre Fernandez, Daniel Haziza, Francisco Massa, Alaaeldin El-Nouby, et~al.
\newblock Dinov2: Learning robust visual features without supervision.
\newblock \emph{arXiv preprint arXiv:2304.07193}, 2023.

\bibitem[Peebles and Xie(2023)]{dit}
William Peebles and Saining Xie.
\newblock Scalable diffusion models with transformers.
\newblock In \emph{Proceedings of the IEEE/CVF International Conference on Computer Vision}, pages 4195--4205, 2023.

\bibitem[Rissanen et~al.(2022)Rissanen, Heinonen, and Solin]{heat}
Severi Rissanen, Markus Heinonen, and Arno Solin.
\newblock Generative modelling with inverse heat dissipation.
\newblock \emph{arXiv preprint arXiv:2206.13397}, 2022.

\bibitem[Rombach et~al.(2022)Rombach, Blattmann, Lorenz, Esser, and Ommer]{ldm}
Robin Rombach, Andreas Blattmann, Dominik Lorenz, Patrick Esser, and Bj{\"o}rn Ommer.
\newblock High-resolution image synthesis with latent diffusion models.
\newblock In \emph{Proceedings of the IEEE/CVF conference on computer vision and pattern recognition}, pages 10684--10695, 2022.

\bibitem[Salimans et~al.(2016)Salimans, Goodfellow, Zaremba, Cheung, Radford, and Chen]{is}
Tim Salimans, Ian Goodfellow, Wojciech Zaremba, Vicki Cheung, Alec Radford, and Xi Chen.
\newblock Improved techniques for training gans.
\newblock \emph{Advances in neural information processing systems}, 29, 2016.

\bibitem[Sauer et~al.(2022)Sauer, Schwarz, and Geiger]{styleganxl}
Axel Sauer, Katja Schwarz, and Andreas Geiger.
\newblock Stylegan-xl: Scaling stylegan to large diverse datasets.
\newblock In \emph{ACM SIGGRAPH 2022 conference proceedings}, pages 1--10, 2022.

\bibitem[Song et~al.(2020)Song, Sohl-Dickstein, Kingma, Kumar, Ermon, and Poole]{vp}
Yang Song, Jascha Sohl-Dickstein, Diederik~P Kingma, Abhishek Kumar, Stefano Ermon, and Ben Poole.
\newblock Score-based generative modeling through stochastic differential equations.
\newblock \emph{arXiv preprint arXiv:2011.13456}, 2020.

\bibitem[Su et~al.(2024)Su, Ahmed, Lu, Pan, Bo, and Liu]{rope}
Jianlin Su, Murtadha Ahmed, Yu Lu, Shengfeng Pan, Wen Bo, and Yunfeng Liu.
\newblock Roformer: Enhanced transformer with rotary position embedding.
\newblock \emph{Neurocomputing}, 568:\penalty0 127063, 2024.

\bibitem[Sun et~al.(2024)Sun, Jiang, Chen, Zhang, Peng, Luo, and Yuan]{llamagen}
Peize Sun, Yi Jiang, Shoufa Chen, Shilong Zhang, Bingyue Peng, Ping Luo, and Zehuan Yuan.
\newblock Autoregressive model beats diffusion: Llama for scalable image generation.
\newblock \emph{arXiv preprint arXiv:2406.06525}, 2024.

\bibitem[Teng et~al.(2023)Teng, Zheng, Ding, Hong, Wangni, Yang, and Tang]{rdm}
Jiayan Teng, Wendi Zheng, Ming Ding, Wenyi Hong, Jianqiao Wangni, Zhuoyi Yang, and Jie Tang.
\newblock Relay diffusion: Unifying diffusion process across resolutions for image synthesis.
\newblock \emph{arXiv preprint arXiv:2309.03350}, 2023.

\bibitem[Teng et~al.(2024)Teng, Wu, Shi, Ning, Dai, Wang, Li, and Liu]{dim}
Yao Teng, Yue Wu, Han Shi, Xuefei Ning, Guohao Dai, Yu Wang, Zhenguo Li, and Xihui Liu.
\newblock Dim: Diffusion mamba for efficient high-resolution image synthesis.
\newblock \emph{arXiv preprint arXiv:2405.14224}, 2024.

\bibitem[Touvron et~al.(2023{\natexlab{a}})Touvron, Lavril, Izacard, Martinet, Lachaux, Lacroix, Rozi{\`e}re, Goyal, Hambro, Azhar, et~al.]{llama1}
Hugo Touvron, Thibaut Lavril, Gautier Izacard, Xavier Martinet, Marie-Anne Lachaux, Timoth{\'e}e Lacroix, Baptiste Rozi{\`e}re, Naman Goyal, Eric Hambro, Faisal Azhar, et~al.
\newblock Llama: Open and efficient foundation language models.
\newblock \emph{arXiv preprint arXiv:2302.13971}, 2023{\natexlab{a}}.

\bibitem[Touvron et~al.(2023{\natexlab{b}})Touvron, Martin, Stone, Albert, Almahairi, Babaei, Bashlykov, Batra, Bhargava, Bhosale, et~al.]{llama2}
Hugo Touvron, Louis Martin, Kevin Stone, Peter Albert, Amjad Almahairi, Yasmine Babaei, Nikolay Bashlykov, Soumya Batra, Prajjwal Bhargava, Shruti Bhosale, et~al.
\newblock Llama 2: Open foundation and fine-tuned chat models.
\newblock \emph{arXiv preprint arXiv:2307.09288}, 2023{\natexlab{b}}.

\bibitem[Wang et~al.(2024)Wang, Li, Song, Li, Ge, Zheng, and Wang]{flowdcn}
Shuai Wang, Zexian Li, Tianhui Song, Xubin Li, Tiezheng Ge, Bo Zheng, and Limin Wang.
\newblock Flowdcn: Exploring dcn-like architectures for fast image generation with arbitrary resolution.
\newblock \emph{arXiv preprint arXiv:2410.22655}, 2024.

\bibitem[Yan et~al.(2023)Yan, Gu, and Rush]{diffusionssm}
Jing~Nathan Yan, Jiatao Gu, and Alexander~M Rush.
\newblock Diffusion models without attention.
\newblock \emph{arXiv preprint arXiv:2311.18257}, 2023.

\bibitem[Yao and Wang(2025)]{vavae}
Jingfeng Yao and Xinggang Wang.
\newblock Reconstruction vs. generation: Taming optimization dilemma in latent diffusion models.
\newblock \emph{arXiv preprint arXiv:2501.01423}, 2025.

\bibitem[Yu et~al.(2024{\natexlab{a}})Yu, He, Deng, Shen, and Chen]{rar}
Qihang Yu, Ju He, Xueqing Deng, Xiaohui Shen, and Liang-Chieh Chen.
\newblock Randomized autoregressive visual generation.
\newblock \emph{arXiv preprint arXiv:2411.00776}, 2024{\natexlab{a}}.

\bibitem[Yu et~al.(2024{\natexlab{b}})Yu, Kwak, Jang, Jeong, Huang, Shin, and Xie]{repa}
Sihyun Yu, Sangkyung Kwak, Huiwon Jang, Jongheon Jeong, Jonathan Huang, Jinwoo Shin, and Saining Xie.
\newblock Representation alignment for generation: Training diffusion transformers is easier than you think.
\newblock \emph{arXiv preprint arXiv:2410.06940}, 2024{\natexlab{b}}.

\bibitem[Yue et~al.(2024)Yue, Wang, Lu, Sun, Wei, Ouyang, Bai, and Zhou]{dod}
Xiaoyu Yue, Zidong Wang, Zeyu Lu, Shuyang Sun, Meng Wei, Wanli Ouyang, Lei Bai, and Luping Zhou.
\newblock Diffusion models need visual priors for image generation.
\newblock \emph{arXiv preprint arXiv:2410.08531}, 2024.

\bibitem[Zhuo et~al.(2024)Zhuo, Du, Xiao, Li, Liu, Huang, Liu, Zhao, Wang, Ma, et~al.]{lumina}
Le Zhuo, Ruoyi Du, Han Xiao, Yangguang Li, Dongyang Liu, Rongjie Huang, Wenze Liu, Lirui Zhao, Fu-Yun Wang, Zhanyu Ma, et~al.
\newblock Lumina-next: Making lumina-t2x stronger and faster with next-dit.
\newblock \emph{arXiv preprint arXiv:2406.18583}, 2024.

\end{thebibliography}
}

\newpage
\appendix
\section{Model Specs}
\begin{table}[h]
    \centering
    \begin{tabular}{l c c c}
    \toprule
     Config & \#Layers & Hidden dim & \#Heads  \\
     \midrule
     B/2 & 12 & 768 & 12 \\
     L/2 & 24 & 1024 & 16 \\
     XL/2 & 28 & 1152 & 16 \\
     \bottomrule
     & 
    \end{tabular}
\end{table}
\section{Hyper-parameters}

\begin{table}[h]
\centering
    \begin{tabular}{c|c}
         \toprule
        VAE & SD-VAE-f8d4-ft-ema \\
        VAE donwsample& 8 \\
        latent channel & 4 \\
        \midrule
        optimizer & AdamW~\cite{adam} \\
        base learning rate & 1e-4 \\
        weight decay & 0.0 \\
        batch size & 256 \\
        learning rate schedule & constant \\
        augmentation & center crop \\
        \midrule
        diffusion sampler & Euler-ODE \\
        diffusion steps & 250 \\
        evaluation suite & ADM~\cite{adm} \\
        \bottomrule
    \end{tabular}
\end{table}

\section{Linear flow and Diffusion}
Given the SDE forward and reverse process:
\begin{align}
     {d}{\bs x}_t &= f(t){\bs x}_t \mathrm{d}t + g(t) \mathrm{d}{\bs w} \\
     {d}{\bs x}_t &= [f(t){\bs x}_t - g(t)^2\nabla_{\bs x} \log p({\bs x}_t)] dt + g(t) {d}{\bs w} 
\end{align}
A corresponding deterministic process exists with trajectories sharing the same marginal probability densities of reverse SDE.
\begin{equation}
     {d}{\bs x}_t = [f(t){\bs x}_t - \frac{1}{2}g(t)^2\nabla_{\bs x_t} \log p({\bs x}_t)] {d}t
\end{equation}
Given $x_t = \alpha_t x_{data} + \sigma \epsilon$. The traditional diffusion model learns:
\begin{equation}
    \nabla_{\bs x_t} \log p({\bs x}_t) = -{\frac{\epsilon}{\sigma(t)}}
\end{equation}
The flow-matching framework actually learns the following: 
\begin{align}
    {\bs v}_t &= \dot{\alpha} x + \dot{\sigma} \epsilon \\
              &= x - \epsilon
\end{align}
    
Here we will demonstrate in flow-matching, the ${\bs v}_t$ prediction is actually as same as the reverse ode:
\begin{align}
& \dot{\alpha} x + \dot{\sigma} \epsilon  \\
=& f(t){\bs x}_t - \frac{1}{2}g(t)^2\nabla_{\bs x_t} \log p({\bs x}_t) 
\label{eq:flow_dif_eq}
\end{align}

Let us start by expanding the reverse ode first.
\begin{align}
     & f(t){\bs x}_t - \frac{1}{2}g(t)^2\nabla_{\bs x_t} \log p({\bs x}_t) \\
    =& f(t)(\alpha(t) {\bs x}_{data} + \sigma(t) \epsilon) - \frac{1}{2}g(t)^2 [{-\frac{\epsilon}{\sigma(t)}}] \\
    =& f(t) \alpha(t) {\bs x}_{data} + (f(t)\sigma(t) + \frac{1}{2} {\frac{g(t)^2}{\sigma(t)}}) \epsilon
\end{align}

To prove \cref{eq:flow_dif_eq}, we needs to demonstrate that:
\begin{align}
    \dot \alpha(t) &= f_t \alpha(t) \label{eq:alpha} \\
    \dot \sigma(t) &= f_t \sigma(t) + \frac{1}{2}\frac{{g_t^2}}{\sigma(t)}. \label{eq:sigma}
\end{align}

Here, let us derive the relation between $f_t$ and $\alpha(t), \dot \alpha(t)$. 
We donate $x_{data}(t) = \alpha(t) x_{data}$ is the remain component of $x_{data}$ in $x_t$, it is easy to find that:
\begin{align}
    d {\bs x}_{data}(t) &= f_t {\bs x}_{data}(t) dt \\
    d (\alpha(t) x_{data}) &= f_t \alpha(t) x_{data} dt \\
    d \alpha(t) &= f_t \alpha(t)  dt 
\end{align}
So, \cref{eq:alpha} is right.

Based on the above equation, we will demonstrate the relation of $g_t, f_t$ with $\sigma(t)$. Note that Gaussian noise has nice additive properties. 
\begin{equation}
    a\epsilon_1 + b\epsilon_2 \in \mathcal{N}(0, \sqrt{a^2+b^2})
\end{equation}
Let us start with the gaussian noise component $\epsilon(t)$ calculation, reaching at $t$, every noise addition at $s \in [0, t]$ while been decayed by a factor of $\frac{\alpha(t)}{\alpha(s)}$. Thus, the mixed Gaussian noise will have a std variance $\sigma(t)$ of:
\begin{align}
    \sigma(t) &= \sqrt{(\int_0^t [(\frac{\alpha(t)}{\alpha(s)})^2 g^2_s] ds)} \\
    \sigma(t) &= \alpha(t)\sqrt{(\int_0^t [(\frac{g_s}{\alpha(s)})^2] ds)}
\end{align}
After obtaining the relation of ${f_t, g_t}$ and ${\alpha(t), \sigma(t)}$, we derive $\dot \alpha(t)$ and $\dot \sigma(t)$ with above conditions:
\begin{align}
   \dot \alpha(t) &= f_t \exp[\int^t_0 f_s ds] \\
   \dot \alpha(t) &= f_t \alpha(t)
\end{align}
As for $\dot \sigma(t)$, it is quit complex but not hard:
\begin{align}
    \dot \sigma(t) &= \dot \alpha(t)\sqrt{(\int_0^t [(\frac{g_t}{\alpha(s)})^2] ds)} + \alpha(t)\frac{\frac{1}{2}\frac{g_t^2}{\alpha(t)}}{\sqrt{(\int_0^t [(\frac{g_t}{\alpha(s)})^2 g^2_s] ds)}} \\
    \dot \sigma(t) &= (f_t\alpha(t)) \sqrt{(\int_0^t [(\frac{g_t}{\alpha(s)})^2] ds)} + \alpha(t)\frac{\frac{1}{2}\frac{g^2_t}{\alpha^2(t)}}{\sqrt{(\int_0^t [(\frac{g_t}{\alpha(s)})^2] ds)}} \\
    \dot \sigma(t) &= f_t \alpha(t) \sqrt{(\int_0^t [(\frac{g_t}{\alpha(s)})^2] ds)}+ \frac{\frac{1}{2}g_t^2}{\alpha(t) \sqrt{(\int_0^t [(\frac{g_t}{\alpha(s)})^2] ds)}} \\
    \dot \sigma(t) &= f_t \sigma(t) + \frac{1}{2}\frac{gt}{\sigma(t)}
\end{align}
So, \cref{eq:sigma} is right.

\section{Proof of Spectrum Autoregressive}
Given the noise scheduler$\{ \alpha_t, \sigma_t\}$, the clean data ${\bs x}_\text{data}$ and Gaussian noise $\epsilon$. Denote $K_{freq}$ as the maximum frequency of the clean data ${\bs x}_{data}$ The noisy latent $x_t$ at timestep $t$ has been defined as:
\begin{equation}
    {\bs x}_t = \alpha_t {\bs x}_{data} + \sigma_t {\bs \epsilon}
\end{equation}

The spectrum magnitude ${\bs c}_i$of $x_t$ on DCT basics ${\bs u}_i$ follows:
\begin{align*}
    {\bs c}_i &=  \mathbb{E}_{\epsilon}[{\bs u}_i^T {\bs x}_t]^2 \\
    {\bs c}_i &=  \mathbb{E}_{\epsilon}[{\bs u}_i^T (\alpha_t {\bs x}_{data} + \sigma_t {\bs \epsilon})]^2
\end{align*}

Recall that the spectrum magnitude of Gaussian noise $\epsilon$ is uniformly distributed.

\begin{align*}
       {\bs c}_i &=  [\alpha_t{\bs u}_i^T  {\bs x}_{data}]^2 + 2\alpha_t\sigma_t\mathbb{E}_{\epsilon}[{\bs u}_i^T {\bs x}_{data} {\bs u}_i^T\epsilon] + \sigma_t^2  \mathbb{E}_{\bs \epsilon}[{\bs u}_i^T \epsilon ]^2 \\
       {\bs c}_i &=  [\alpha_t{\bs u}_i^T  {\bs x}_{data}]^2 + \sigma_t^2  \mathbb{E}_{\bs \epsilon}[{\bs u}_i^T \epsilon ]^2 \\
      {\bs c}_i &=  \alpha_t^2[{\bs u}_i^T  {\bs x}_{data}]^2 + \sigma_t^2 \lambda 
\end{align*}

if $\sigma_t^2 \lambda$ has bigger value than $[\alpha_t{\bs u}_i^T  {\bs x}_{data}]^2$, the spectrum magnitude ${\bs c}_i$ on DCT basics ${\bs u}_i$ will be canceled, thus the maximal remaining frequency $f_{max}(t)$ of original data in ${\bs x}_t$ follows:
\begin{equation}
    f_{max}(t) > \min\left({\left(\frac{\alpha_t{\bs u}_i^T  {\bs x}_{data}}{\sigma_t \lambda}\right)}^2, K_{freq}\right)
\end{equation}

Though ${\frac{\alpha_t{\bs u}_i^T {\bs x}_{data}}{\sigma_t \lambda}}^2$ depends on the dataset. Here, we directly suppose it as a constant $1$. And replace $\alpha = t$ and $\sigma=1-t$ in above equation:
\begin{equation}
f_{max}(t) > \min\left({\left(\frac{t}{1-t}\right)}^2, K_{freq}\right)
\end{equation}

\section{Linear multisteps method}
We conduct targeted experiment on SiT-XL/2 with Adams–Bashforth like linear multistep solver; To clarify, we did not employ this powerful solver for our DDT models in all tables across the main paper.

The reverse ode of the diffusion models tackles the following integral:
\begin{equation}
     {\bs x}_{i+1} = {\bs x}_i + \int_{t_i}^{t_{i+1}} {\bs v}_\theta({\bs x}_t, t) {dt} \label{eq:fm_split_int} \\
\end{equation}
The classic Euler method employs ${\bs v}_\theta({\bs x}_i, t_i)$ as an estimate of ${\bs v}_\theta({\bs x}_t, t)$ throughout the interval $[t_i, t_{i+1}]$
\begin{equation}
     {\bs x}_{i+1} = {\bs x}_i + (t_{i+1} -t_{i}) {\bs v}_\theta({\bs x}_i, t_i). \\
\end{equation}
The most classic multi-step solver Adams–Bashforth method~(deemed as Adams for brevity) incorporates the Lagrange polynomial to improve the estimation accuracy with previous predictions. 
\begin{align*}
     {\bs v}_\theta({\bs x}_t, t)  &=  \sum_{j=0}^i (\prod_{k=0,k\neq j}^i{\frac{t-t_k}{t_j - t_k}}){\bs v}_\theta({\bs x}_j, t_j) \\
    {\bs x}_{i+1} &\approx {\bs x}_i + \int_{t_i}^{t_{i+1}} \sum_{j=0}^i (\prod_{k=0,k\neq j}^i{\frac{t-t_k}{t_j - t_k}}){\bs v}_\theta({\bs x}_j, t_j) dt  \\
     {\bs x}_{i+1} &\approx {\bs x}_i + \sum_{j=0}^i {\bs v}_\theta({\bs x}_j, t_j) \int_{t_i}^{t_{i+1}} (\prod_{k=0,k\neq j}^i{\frac{t-t_k}{t_j - t_k}}) dt 
\end{align*}
Note that $\int_{t_i}^{t_{i+1}} (\prod_{k=0,k\neq j}^i{\frac{t-t_k}{t_j - t_k}}) dt$ of the Lagrange polynomial can be pre-integrated into a constant coefficient, resulting in only naive summation being required for ODE solving.

\section{Classifier free guidance.} 
As classifier-free guidance significantly impacts the performance of diffusion models. Traditional classifier-free guidance improves performance at the cost of decreased diversity. Interval guidance is recently been adopted by REPA\cite{repa} and Causalfusion\cite{causalfusion}, It applies classifier-free guidance only to the high-frequency generation phase to preserve the diversity. We sweep different classifier-free guidance strength with selected intervals. Our DDT-XL/2 achieves the best performance with interval $[0.3, 1]$ with a classifer-free guidance of 2. Recall that we donate $t=0$ as the pure noise timestep while REPA\cite{repa} use $t=1$, thus this exactly correspond to the $[0, 0.7]$ interval in REPA\cite{repa}

\begin{figure}
    \centering
    \includegraphics[width=\linewidth]{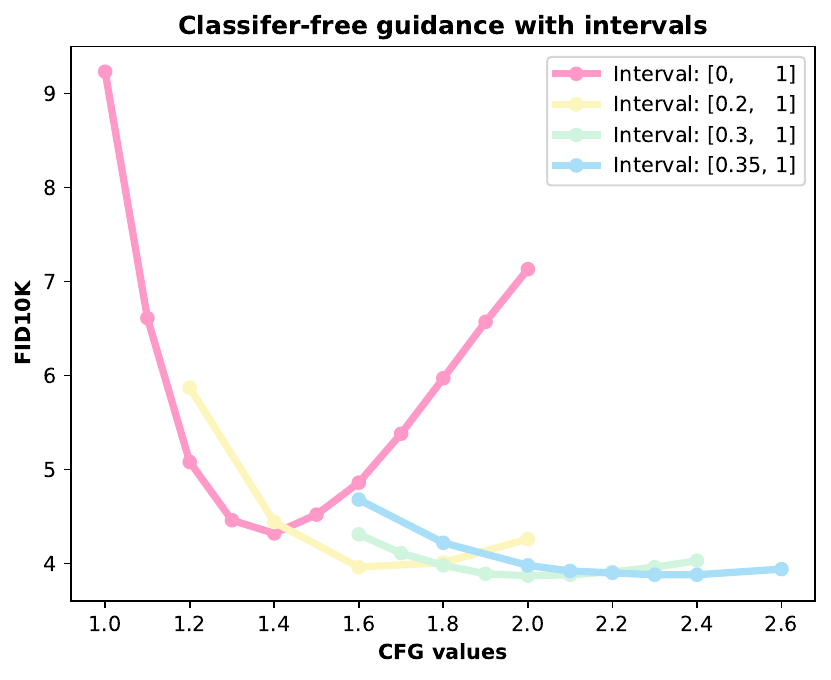}
    \caption{\textbf{FID10K of DDT-XL/2 with different Classifer free guidance strength and guidance intervals.} {\small We sweep different classifier-free guidance strength with selected intervals. Our DDT-XL/2 achieves the best performance with interval $[0.3, 1]$ with a classifer-free guidance of 2.}}
    \label{fig:cfg}
    \vspace{-1em}
\end{figure}

\end{document}